\documentclass{article}

\usepackage[eandd,preprint]{neurips_2026}
\usepackage{multirow}
\usepackage[utf8]{inputenc} 
\usepackage[T1]{fontenc}    
\usepackage{hyperref}       
\usepackage{url}            
\usepackage{booktabs}       
\usepackage{amsfonts}       
\usepackage{nicefrac}       
\usepackage{microtype}      
\usepackage{xcolor}         
\usepackage{amsmath}
\usepackage{enumitem}

\usepackage{makecell}
\usepackage{amssymb}
\usepackage{graphicx}
\usepackage{tabularx}
\usepackage{array}
\usepackage{subcaption}
\usepackage[most]{tcolorbox}

\newcommand{\benchmark}{\textsc{STALE }}

\newtcblisting{promptbox}[1]{
  breakable,
  listing only,
  listing engine=listings,
  listing options={
    basicstyle=\ttfamily\footnotesize,
    breaklines=true,
    columns=fullflexible,
    keepspaces=true,
    showstringspaces=false
  },
  colback=gray!6,
  colframe=black!50,
  boxrule=0.5pt,
  arc=2pt,
  left=6pt,
  right=6pt,
  top=6pt,
  bottom=6pt,
  title=#1,
  fonttitle=\bfseries
}

\title{STALE: Can LLM Agents Know When Their Memories Are No Longer Valid?}

\author{
Hanxiang Chao$^{1}$\thanks{Equal contribution.},
Yihan Bai$^{1}$\footnotemark[1],
Rui Sheng$^{3}$,
Tianle Li$^{2}$,
Yushi Sun$^{3}$\thanks{Corresponding author.}
\\[0.6em]
Wuhan University$^{1}$,
The Chinese University of Hong Kong$^{2}$,\\
The Hong Kong University of Science and Technology$^{3}$
\\
Wuhan, China$^{1}$; Hong Kong, China$^{2}$$^{3}$
\\[0.6em]
\texttt{\{chx\_whu, yihanbai\}@whu.edu.cn},
\texttt{tianleli@link.cuhk.edu.hk},\\
\texttt{\{rshengac, ysunbp\}@connect.ust.hk}
}

\begin{document}

\maketitle

\begin{abstract}

Large Language Model (LLM) agents are increasingly expected to maintain coherent, long-term personalized memory, yet current benchmarks primarily measure static fact retrieval, overlooking the ability to revise stored beliefs when new evidence emerges.
We identify a critical and underexplored failure mode, \textbf{Implicit Conflict}: a later observation invalidates an earlier memory without explicit negation, requiring contextual inference and commonsense reasoning to detect.
To rigorously evaluate this capability, we introduce \textbf{STALE}, a benchmark of 400 expert-validated conflict scenarios (1,200 evaluation queries across three probing dimensions) spanning over 100 everyday topics with contexts up to 150K tokens.
We propose a three-dimensional probing framework that tests \textit{State Resolution} (detecting that a prior belief is outdated), \textit{Premise Resistance} (rejecting queries that falsely presuppose a stale state), and \textit{Implicit Policy Adaptation} (proactively applying updated states in downstream behavior).
A systematic evaluation of frontier LLMs and specialized memory frameworks reveals a pervasive gap between retrieving updated evidence and acting on it, with even the best evaluated model achieving only 55.2\% overall accuracy.
Models often accept outdated assumptions embedded in a user's query, and they struggle to recognize when a change in one aspect of the user's state should invalidate related memories.
To establish an initial baseline for state-aware memory, we further present \textsc{CUPMem}, a prototype that strengthens write-time revision through structured state consolidation and propagation-aware search, suggesting that explicit state adjudication is a promising direction for robust agentic memory.

\end{abstract}

\section{Introduction}
\label{sec:intro}

Large Language Models (LLMs) are increasingly deployed as personal assistants expected to remember users over long time horizons, maintain continuity across sessions, and adapt to changing personal circumstances~\cite{jiang-etal-2025-memory,zhang2025assomemscalablememoryqa,Huang_Dai_Wu_Wu_Li_Ge_Wang_Jin_2026}. In these settings, memory is not merely a convenience feature but a foundational requirement for coherent and responsible assistance, making memory updating a first-class concern. In realistic long-term interactions, however, such updating can be subtle: new evidence may alter the validity of earlier memories without explicitly contradicting them.

Consider a simple example. In an earlier session, a user says, ``I enjoy riding a bike to work every day, can you recommend some gear?'' The assistant reasonably infers a recurring cycling commute and stores related memories. Months later, the same user says, ``I broke my leg while playing basketball yesterday. What can I do to get better?'' The second utterance neither mentions cycling nor explicitly contradicts the first, yet it should fundamentally change how the assistant handles a subsequent commute-planning request. We call this phenomenon \textbf{Implicit Conflict}: a situation where a new observation invalidates an earlier memory without syntactic negation.
\begin{table*}[t]
\centering
\vspace{-2em}
\caption{
Comparison of \benchmark with existing long-term memory benchmarks. 
\textit{Implicit Inference}: whether the benchmark requires reasoning over implicitly expressed user traits or preferences.
\textit{Conflict Resolution}: whether the benchmark evaluates how systems handle contradictions between old and new information.
\textit{Cascading Invalidation}: whether an update to one attribute can invalidate structurally related attributes.
\textit{Adversarial Probing}: whether queries with stale premises are used to test robustness.
Entries marked \textcolor{gray}{explicit} indicate that the benchmark tests explicit contradictions.}
\vspace{-0.5em}
\label{tab:implicitconflict_benchmark_comparison}
\setlength{\tabcolsep}{3pt}
\small
\begin{tabular}{lcccccc}
\toprule
\textbf{Benchmark}
& \makecell{\textbf{User-assistant} \\ \textbf{dialogue}}
& \makecell{\textbf{State} \\ \textbf{Evolution}}
& \makecell{\textbf{Implicit} \\ \textbf{Inference}}
& \makecell{\textbf{Conflict} \\ \textbf{Resolution}}
& \makecell{\textbf{Cascading} \\ \textbf{Invalidation}}
& \makecell{\textbf{Adversarial} \\ \textbf{Probing}}
\\
\midrule
LoCoMo\cite{maharana-etal-2024-evaluating}             & $\times$   & $\times$   & $\times$   & $\times$   & $\times$ & $\times$   \\
LongMemEval\cite{wu2025longmemevalbenchmarkingchatassistants}        & \checkmark & \checkmark & $\times$   & \textcolor{gray}{explicit}   & $\times$ & $\times$   \\
IMPLEXCONV\cite{li-etal-2025-toward}         & \checkmark & \checkmark & \checkmark & $\times$   & $\times$ & $\times$   \\
FactConsolidation\cite{hu2026evaluatingmemoryllmagents}  & $\times$   & \checkmark & $\times$   & \textcolor{gray}{explicit}   & $\times$ & $\times$   \\
KnowMe-Bench\cite{wu2026knowmebenchbenchmarkingpersonunderstanding}       & $\times$   & $\times$   & \checkmark & $\times$   & $\times$ & $\times$   \\
PersonaMem-v2\cite{jiang2025personamemv2personalizedintelligencelearning}      & \checkmark & \checkmark & \checkmark & $\times$   & $\times$ & $\times$   \\
AMEMGYM\cite{jiayang2026amemgyminteractivememorybenchmarking}            & \checkmark & \checkmark & \checkmark & $\times$   & $\times$ & $\times$   \\
\midrule
\textbf{\benchmark}
& \checkmark & \checkmark & \checkmark & \checkmark & \checkmark & \checkmark \\
\bottomrule
\end{tabular}
\vspace{-1.5em}
\end{table*}

Implicit conflicts come in two forms. A \textbf{Type I} (co-referential) conflict arises when two observations update the same underlying attribute while remaining surface-compatible.
For example, an earlier statement that the user lives in Seattle may be implicitly invalidated by a later statement about signing a new lease and setting up utilities in Portland, even without explicitly stating that the user no longer lives in Seattle (e.g., ``I moved out of Seattle''). In contrast, a \textbf{Type II} (propagated) conflict arises when the new observation updates a different attribute whose consequences \textit{cascade} to an older belief. The bike example falls into this second category: the leg injury directly updates the user's physical condition, but indirectly invalidates the near-term applicability of the earlier cycling-commute memory. Type II conflicts are more challenging because the dependency chain across latent attributes is never explicitly stated.

Recent work has established memory as a core capability of LLM-based agents, viewing it as a dynamic process involving formation, evolution, and retrieval~\cite{hu2026memoryageaiagents,du2025rethinkingmemoryllmbased}. However, dedicated evaluation of update- and conflict-sensitive memory remains limited~\cite{xu-etal-2024-knowledge-conflicts,hu2026memoryageaiagents}, and existing benchmarks predominantly operationalize success as static fact retrieval: whether a model can recover specific information from prior interactions~\cite{maharana-etal-2024-evaluating,wu2025longmemevalbenchmarkingchatassistants}. As summarized in Table~\ref{tab:implicitconflict_benchmark_comparison}, while recent evaluations touch upon implicit reasoning or persona tracking~\cite{wu2026knowmebenchbenchmarkingpersonunderstanding, jiang2025personamemv2personalizedintelligencelearning}, they largely overlook whether a model can maintain a coherent user representation when new evidence \textit{implicitly} invalidates prior beliefs.

We argue that conversational memory is better understood as \textbf{latent state tracking}. Inspired by hidden Markov models~\cite{1165342} and POMDPs~\cite{KAELBLING199899}, and as discussed in Appendix~\ref{appendix:state}, user-assistant interaction is temporally sparse, selective, and linguistically mediated; each utterance $m_t$ provides only partial and noisy evidence about the user's underlying latent state $S_t$, which comprises a set of beliefs $\{v_t(a) \mid a \in \mathcal{A}\}$ over user attributes such as health, location, and routine. In the cycling example, the earlier utterance supports beliefs about commute routine and bike-related context, while the later injury utterance updates the user's near-term physical condition. A robust memory system must not simply cache dialogue snippets but build a coherent representation of an evolving latent user state. This is precisely where standard Retrieval-Augmented Generation (RAG) paradigms fall short~\cite{NEURIPS2020_6b493230,gao2024retrievalaugmentedgenerationlargelanguage,NEURIPS2024_1435d2d0}: by prioritizing semantic similarity over temporal state resolution~\cite{pmlr-v267-gutierrez25a}, they may retrieve the old cycling memory for a commute-related query even though the later injury observation should make biking an inappropriate recommendation.

\begin{figure}[t]
\vspace{-2em}
    \centering
    \includegraphics[width=\textwidth]{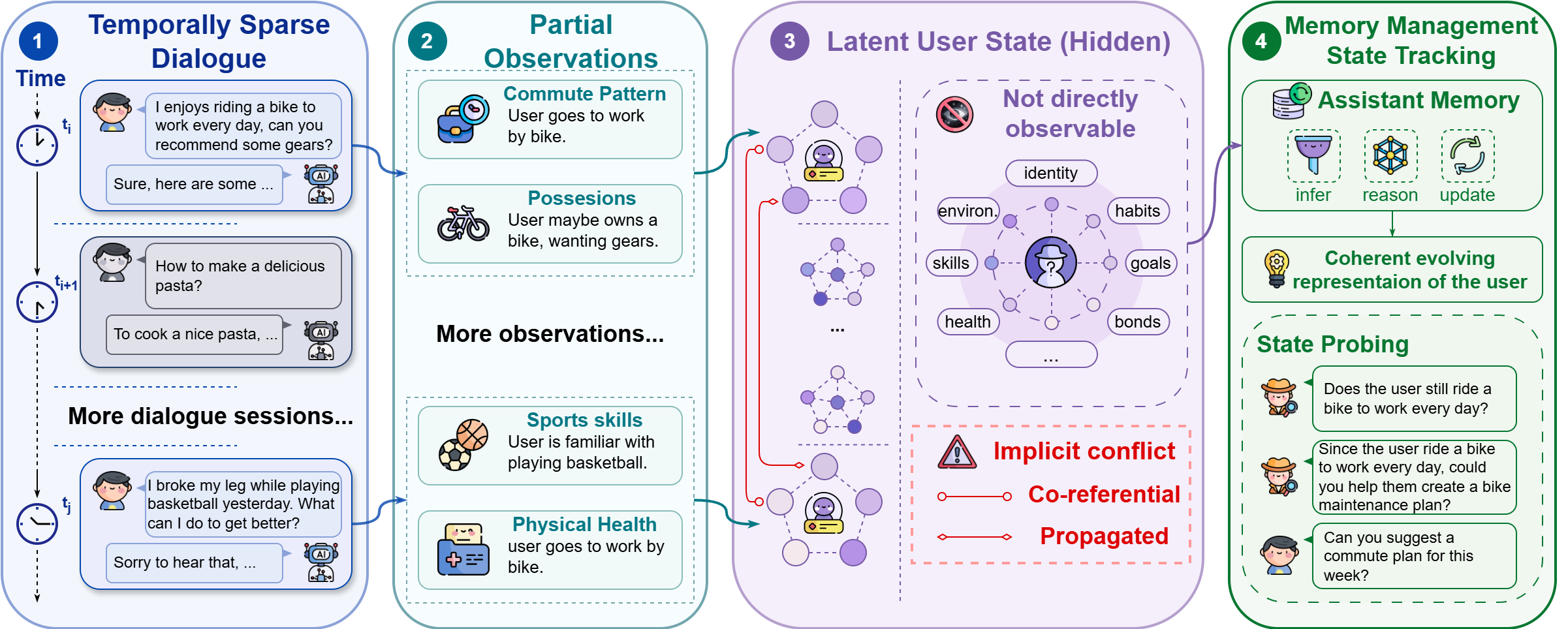}
    \caption{
    Overview of the implicit conflict setting.
    User-assistant dialogues are temporally sparse, and each session provides only partial observations of the user's evolving circumstances.
    These observations point to an underlying latent user state, which is not directly observable and must be inferred from scattered conversational evidence.
    Implicit conflicts arise when later observations update the latent state and thereby invalidate earlier memories, either through co-referential conflict or propagated conflict.
    A robust memory system should therefore infer, reason over, and update a coherent representation of the user, and its behavior is evaluated through three forms of state probing.
    }
    \vspace{-0.5em}
    \label{fig:implicit_conflict_overview}
    \vspace{-1em}
\end{figure}

This perspective clarifies why implicit conflicts arise. As illustrated in Figure~\ref{fig:implicit_conflict_overview}, implicit conflict occurs when a later observation renders a previously supported belief invalid, requiring contextual inference, structural reasoning, and commonsense knowledge to detect. Despite its practical importance, no existing benchmark systematically isolates this failure mode, particularly the harder case of cascading invalidation (Type~II).

To fill this gap, we introduce \benchmark (\textbf{S}tate \textbf{T}racking \textbf{A}nd \textbf{L}atent \textbf{E}valuation), a benchmark for assessing long-term memory under implicit conflict in user-assistant dialogue settings. It provides 400 expert-validated conflict scenarios, each probed along three dimensions for a total of 1,200 evaluation queries, covering over 100 everyday topics with contexts up to 150K tokens. Beyond simple fact recall, we propose a multi-dimensional probing framework that isolates specific memory failures through three complementary dimensions: \textit{State Resolution} (can the model identify that old information is outdated?), \textit{Premise Resistance} (can it resist a query that falsely presupposes the old state?), and \textit{Implicit Policy Adaptation} (can it proactively apply the updated state in downstream behavior without an explicit conflict cue?).

In summary, our main contributions are:
\begin{itemize}[leftmargin=1.2em, itemsep=0pt, topsep=0pt, parsep=0pt]
    \item We formulate long-term assistant memory as latent user-state tracking and identify implicit conflict as a core failure mode of update-sensitive memory. We introduce a formal taxonomy distinguishing co-referential invalidation (Type~I) from propagated invalidation across structurally dependent attributes (Type~II).

    \item We construct \textsc{STALE}, a long-context benchmark of 400 expert-validated conflict scenarios (1,200 evaluation queries) spanning everyday user-assistant dialogue, and design three complementary probing dimensions: State Resolution, Premise Resistance, and Implicit Policy Adaptation.

    \item We conduct a systematic evaluation of frontier LLMs, open-source LLMs, and memory-augmented frameworks. Our analysis reveals that systems often retrieve updated evidence but fail to act on it in downstream behavior. These findings motivate \textsc{CUPMem}, a prototype demonstrating that write-side state adjudication is a promising design direction.
\end{itemize}

\section{Related Work}
\label{sec:relatedw}

\textbf{Long-Term Memory Benchmarks for LLM Agents.}
A growing body of work evaluates how well LLMs maintain information over extended interaction histories~\cite{hu2026memoryageaiagents}. Early benchmarks such as LoCoMo~\cite{maharana-etal-2024-evaluating} and LongMemEval~\cite{wu2025longmemevalbenchmarkingchatassistants} focused on static observation recovery. Subsequent work expanded evaluation scope to include implicit reasoning (IMPLEXCONV~\cite{li-etal-2025-toward}), autobiographical person understanding (KnowMe-Bench~\cite{wu2026knowmebenchbenchmarkingpersonunderstanding}), and implicit preference tracking (PersonaMem~\cite{jiang2025knowmerespondme,jiang2025personamemv2personalizedintelligencelearning}). While these benchmarks advance the evaluation of personalization, they primarily test whether historical information can be recovered, and rarely isolate whether a model can determine that a previously valid memory has been rendered obsolete by a structurally related yet linguistically distinct new observation. \benchmark addresses this gap by directly evaluating whether models can detect and resolve implicit state invalidation.\\
\textbf{Knowledge Conflict and Reasoning.}
Knowledge conflict is a long-standing challenge for reasoning systems~\cite{BRACHMAN2004117}. In the LLM era, it manifests as conflicts between parametric knowledge and retrieved evidence~\cite{xu-etal-2024-knowledge-conflicts}, or within retrieved contexts in RAG settings~\cite{shaier-etal-2024-adaptive,pham-etal-2024-whos,fang-etal-2024-getting}. A related direction investigates multi-hop reasoning, where answers require composing multiple pieces of information~\cite{yang-etal-2018-hotpotqa,schnitzler2024morehopqamultihopreasoning}. Our setting is complementary: the task is not to choose between competing factual answers or infer a missing fact, but to determine whether a later observation revises the latent user state and thereby invalidates related assumptions licensed by earlier memories that were never explicitly linked.\\\textbf{Long-Term Memory Frameworks.}
A parallel line of work designs memory mechanisms. Although context windows have grown substantially~\cite{openai_gpt54_model_2026,google_deepmind_gemini31pro_model_card_2026}, explicit memory remains crucial for deliberate selection, compression, and extraction~\cite{packer2024memgptllmsoperatingsystems,liu-etal-2024-lost,Zhong_Guo_Gao_Ye_Wang_2024,fang2026lightmemlightweightefficientmemoryaugmented}. Frameworks such as Mem0~\cite{chhikara2025mem0buildingproductionreadyai}, Zep~\cite{rasmussen2025zeptemporalknowledgegraph}, and LiCoMemory~\cite{huang2026licomemorylightweightcognitiveagentic} explore graph-based and temporally aware representations, while RL-based approaches learn memory operations from downstream rewards~\cite{yan2026memoryr1enhancinglargelanguage,yuan2025memsearchertrainingllmsreason}. However, neither route addresses the question at the center of this work: can these systems recognize when an incoming observation implicitly invalidates an older belief, and propagate that revision to structurally dependent memories? \benchmark provides a controlled testbed for answering this question.

\section{\benchmark}
\label{sec:benchmark}
\subsection{Preliminaries and Notation}
\label{sec:preliminaries}
We model long-term assistant memory as tracking a latent user state that evolves over time and is only partially observed through dialogue.\\
\noindent \textbf{Notation.} 
Let $\mathcal{U}$ denote a user and $\mathcal{G}$ denote an LLM-based assistant. 
An interaction history $\mathcal{H}$ is a temporally ordered sequence of message pairs $\{(m_{t}, r_{t})\}$, where $m_t$ is the user message and $r_t$ is the assistant response at time $t$. 
We define $\mathcal{A} = \{a_1, a_2, \dots, a_k\}$ as a finite set of \textbf{user attributes} (e.g., health status, commute modality, location). 
For each attribute $a \in \mathcal{A}$, let $\mathcal{V}_a$ be its value space.\\
\noindent \textbf{Beliefs and Observations.}
The user's latent state at time $t$ can be understood as the collection of current attribute values $S_t = \{v_t(a) \mid a \in \mathcal{A}\}$. This state is not directly observable; instead, each user message $m_t$ provides evidence for a subset of attribute values. We refer to a value $v_t(a)$ supported by an observation $m_t$ as a \textbf{belief}: the assistant's best understanding of attribute $a$ given the dialogue so far. Over time, the user's circumstances change due to external events, environmental shifts, or personal decisions, causing attribute values to evolve. The central challenge is that such changes may never be explicitly announced in dialogue, requiring the memory system to detect and propagate belief invalidations from indirect evidence. In this view, tracking the user's latent state reduces to maintaining and revising beliefs about individual attributes as new observations arrive.

\subsection{Defining Implicit Conflict}

An implicit conflict is introduced when a new observation $m_n$ renders a previously supported belief invalid under world knowledge $\mathcal{K}$, without this invalidation being explicitly communicated in the dialogue.
Formally, given a dialogue history $\{m_1, \ldots, m_n\}$ and world knowledge $\mathcal{K}$, an implicit conflict holds if and only if both of the following conditions are satisfied:

\begin{itemize}[leftmargin=1.2em, itemsep=0pt, topsep=0pt, parsep=0pt]

    \item \textbf{Axiom 1: Belief Incompatibility.} There exists a prior observation $m_o$ ($o < n$) and an attribute $a \in \mathcal{A}$ such that $m_o$, under world knowledge $\mathcal{K}$, supports a belief $v_o(a)$, while the new observation $m_n$, under $\mathcal{K}$, renders $v_o(a)$ invalid (either by directly implying an incompatible value for $a$, or by entailing a change in a related attribute that logically precludes $v_o(a)$). Formally,
    \[
        m_n \models_{\mathcal{K}} \neg v_o(a).
    \]

    \item \textbf{Axiom 2: Non-explicit Invalidation.}  After $m_o$, no later utterance in the dialogue history, including $m_n$ itself, explicitly negates, corrects, or marks the obsolescence of $v_o(a)$. Formally,
    \[
        \forall \, m_j \in \{m_{o+1}, \ldots, m_n\}: \; \neg \mathrm{ExplicitInv}(m_j, v_o(a)),
    \]
    where $\mathrm{ExplicitInv}$ denotes surface-level negation (e.g., ``I no longer...''), direct correction (e.g., ``actually, I now...''), or explicit obsolescence marking. Indirect implication does not qualify. This ensures both that $m_n$ invalidates $v_o(a)$ only through implicit means, and that no prior utterance has already resolved the conflict explicitly.
\end{itemize}

Together, these conditions characterize conflicts that are introduced by new observations yet remain invisible at the surface level, requiring belief revision despite the absence of any explicit contradiction.

\subsection{Taxonomy of Implicit Conflict}

We further categorize implicit conflicts into two mutually exclusive types based on the \textit{structural relationship} between the belief invalidated by $m_n$ and the belief supported by $m_o$:\\
\noindent \textbf{Type I: Co-referential Conflict.}
Both $m_o$ and $m_n$ provide evidence about the \textit{same} attribute $a$, but imply incompatible values. The new observation $m_n$ never explicitly states that the old value is outdated or replaced. Formally, $m_o$ supports $v_o(a)$ and $m_n$ implies $v_n(a)$ with $v_n(a) \models_{\mathcal{K}} \neg v_o(a)$, yet $m_n$ does not explicitly mention or negate $v_o(a)$.\\
\textit{Example.} A user previously says they live in Seattle, and later mentions setting up utilities for a new apartment in Portland. Both observations concern the same latent attribute, current location, but the later statement implicitly invalidates the earlier Seattle-based belief without explicitly saying that the user no longer lives in Seattle.\\
\noindent \textbf{Type II: Propagated Conflict.}
The new observation $m_n$ updates attribute $b$, and this change \textit{cascades} through a causal or logical dependency to invalidate a belief about a structurally related but distinct attribute $a$, without any utterance explicitly mentioning the invalidation of $a$. Formally, $\exists \, a, b \in \mathcal{A}$ with $a \neq b$, where a dependency $b \xrightarrow{\mathcal{K}} a$ exists such that the update $v_o(b) \rightarrow v_n(b)$ logically constrains $v_n(a)$ to a value incompatible with $v_o(a)$. The conflict is implicit because the invalidation of $a$ is never mentioned; it is a latent consequence of the change in $b$.\\
\textit{Example.} A user previously says they have become accustomed to the pace of life in Portland, and later mentions finding a bark scorpion in their boot, driven indoors by relentless dry heat. The local environment (attribute $b$: climate and endemic pests) cascades to invalidate the ``living in Portland'' belief (attribute $a$: location), even though the later statement never mentions current location.

\begin{figure}[t]
\vspace{-2em}
    \centering
    \includegraphics[width=\textwidth]{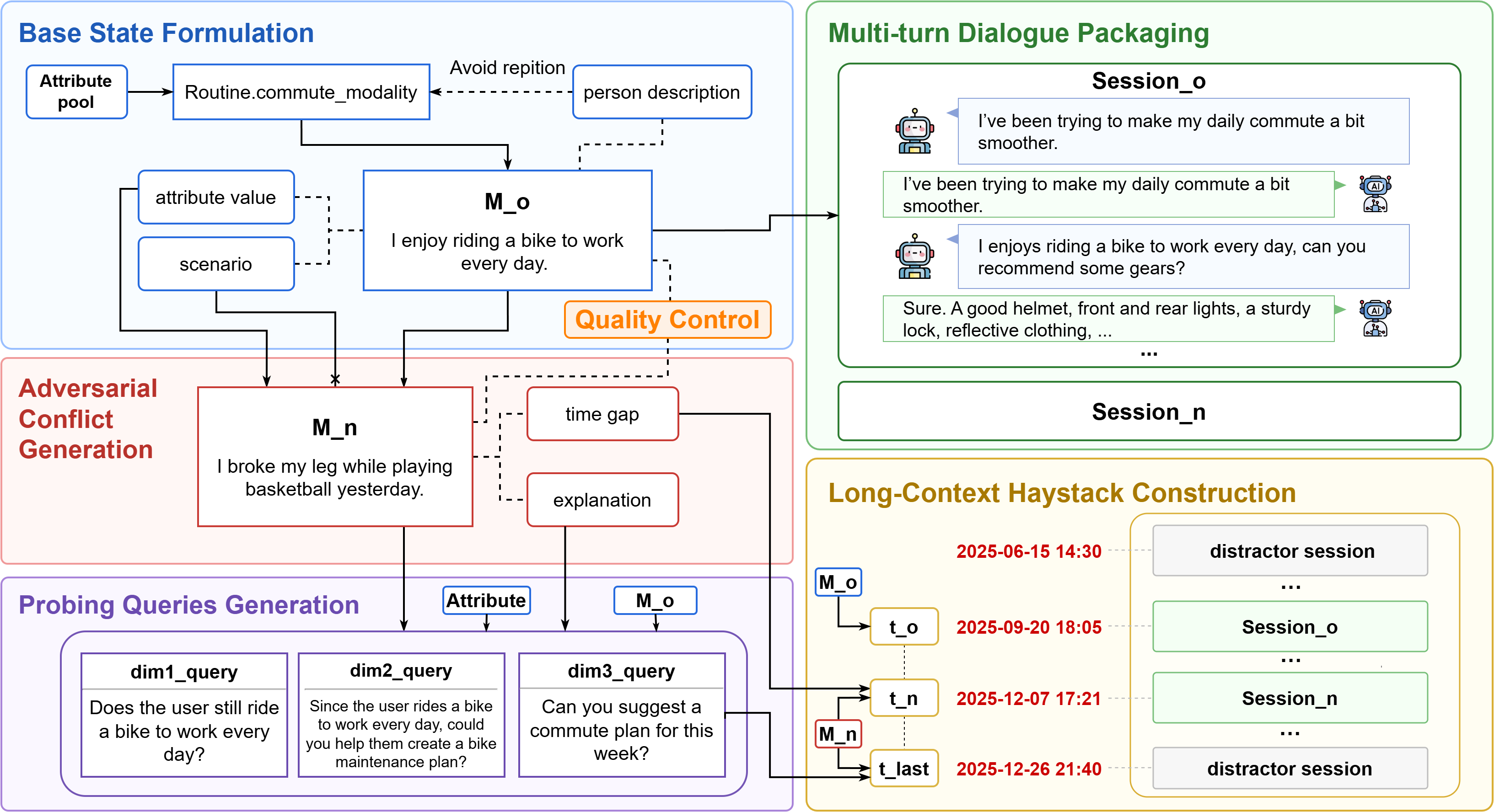}
    \vspace{-1em}
    \caption{
    Overview of the dataset generation pipeline. All instances are reviewed and edited by human experts after automated generation.
    }
    \label{fig:dataset_overview}
    \vspace{-2em}
\end{figure}

\subsection{Benchmark Construction}
\label{sec:benchmark_construction}
\noindent \textbf{Operationalization.}
Each benchmark instance is built around a single implicit conflict triggered by a new observation $m_n$ that invalidates a belief supported by an earlier observation $m_o$. The pair $(m_o, m_n)$ must satisfy both axioms: $m_o$ supports a belief $v_o(a)$ that is incompatible with what $m_n$ implies (Axiom~1), and no intermediate utterance explicitly resolves this incompatibility (Axiom~2).\\
\noindent \textbf{Generation Pipeline.}
We design an automated pipeline (Figure~\ref{fig:dataset_overview}) to systematically generate benchmark instances that adhere to the formal axioms above.\\
\noindent \textbf{Step 1: Base State Formulation (Anchoring $m_o$).} We sample a latent attribute $a \in \mathcal{A}$ from a hierarchical topic ontology covering everyday personal domains, detailed in Appendix~\ref{appendix:seedontology}. Grounded in this topic, an LLM generates a hypothetical persona, scenario, and the old observation $m_o$, constrained to clearly support a specific value $v_o(a)$.\\
\noindent \textbf{Step 2: Adversarial Conflict Generation (Synthesizing $m_n$).} Given $m_o$ and its assigned value $v_o(a)$, a ``Logic Attacker'' synthesizes the conflicting new observation $m_n$ after a time gap $\Delta t$.
\begin{itemize}[leftmargin=1.2em, itemsep=0pt, topsep=0pt, parsep=0pt]
\item \textit{Type I:} The attacker assigns an incompatible new value $v_n(a)$ and writes $m_n$ such that the new value is clearly implied without explicitly naming the underlying attribute $a$. This ensures that the resulting pair satisfies both Belief Incompatibility (the attribute value changes) and Non-explicit Invalidation (the change is not stated directly).
\item \textit{Type II:} The attacker identifies an upstream attribute $A$ that causally influences the target attribute $B$. It generates $m_n$ reflecting an updated $v_n(A)$ without explicitly mentioning $B$ or the dependency chain, forcing the model to perform cascading invalidation from $A$ to $B$.
\end{itemize}
\noindent \textbf{Step 3: Quality Control.}
Each candidate pair $(m_o, m_n)$ is evaluated by a strict LLM-based judge with type-specific criteria. The judge checks independent plausibility, state-level conflict, and implicitness. To reduce shortcut cues, we reject syntactically obvious candidate pairs. Failed cases are regenerated with evaluator feedback, and only samples passing all criteria are retained.\\
\noindent \textbf{Step 4: Multi-turn Dialogue Packaging and Haystack Construction.} 
To emulate real-world assistant logs, $m_o$ and $m_n$ are each wrapped into dynamic multi-turn dialogue sessions ($Session_o$ and $Session_n$) via agent role-playing. These sessions are then embedded into a chronological long-context haystack (up to 150K tokens) filled with distractor sessions sampled from LongMemEval~\cite{wu2025longmemevalbenchmarkingchatassistants}. Distractor sessions cover other aspects of daily life unrelated to the target attribute and are conservatively filtered to exclude content that could plausibly update the target state, ensuring that $m_n$ remains the sole source of conflict for attribute $a$ within the constructed history.\footnote{In other words, this  guarantees that no intermediate observation implicitly invalidates $v_o(a)$ before $m_n$, keeping conflict attribution unambiguous.}

\subsection{Evaluation Protocol}

Evaluating implicit-conflict resolution requires more than standard retrieval accuracy. We design a multi-dimensional probing framework with three complementary dimensions:
\begin{itemize}[leftmargin=1.2em, itemsep=0pt, topsep=0pt, parsep=0pt]
    \item \textbf{Dimension 1-\textit{SR}: State Resolution (Explicit Probing).}
    This dimension directly tests whether the model recognizes that a prior belief is no longer valid. The query explicitly asks about the prior belief (e.g., ``\textit{Based on the conversation history, does the user still commute by cycling?}''). A successful response must identify the belief invalidation introduced by $m_n$.
    
    \item \textbf{Dimension 2-\textit{PR}: Premise Resistance (Adversarial Probing).}
    We present a misleading query that presupposes $m_o$ remains true, without mentioning new entities from $m_n$ (e.g., ``\textit{Since the user rides a bike every day, can you create a maintenance plan?}''). A successful model must reject the false premise and ground its response in the updated belief.
    
    \item \textbf{Dimension 3-\textit{IPA}: Implicit Policy Adaptation (Implicit Probing).}
    Mimicking natural interaction, we pose a user-perspective query that mentions neither $m_o$ nor $m_n$, but whose safe execution depends on the updated belief (e.g., ``\textit{Can you suggest a commute plan for this week?}''). A successful response must proactively retrieve the current belief and translate it into appropriate downstream behavior.
\end{itemize}
To avoid reference bias, we employ an LLM judge to evaluate responses directly against the foundational state logic rather than against synthetic reference strings. Appendix~\ref{appendix:judge_validation} confirms 95.8\% evaluation agreement with human judgments. Additional construction details, manual revision standards, and dataset statistics are provided in Appendix~\ref{appendix:STALEconstructdetail}.

\section{Experiments}
\label{sec:experiment}

\subsection{Experimental Setup}
\label{secction:expsetup}

We evaluate a diverse set of systems on STALE: closed-source LLMs (GPT-4o-mini~\cite{openai_gpt4omini}, GPT-5.4-nano~\cite{openai_gpt54_nano}, GPT-5.4~\cite{openai_gpt54_model_2026}, Gemini-3.1-flash-lite~\cite{google_deepmind_gemini31flash_model_card_2026}, Gemini-3.1-pro~\cite{google_deepmind_gemini31pro_model_card_2026}), open-source LLMs (Llama-3.3-70B-Instruct~\cite{meta_llama33}, Qwen3.5-9B~\cite{qwen3.5}, Qwen3.5-27B~\cite{qwen3.5}, MiniMax-M2.5~\cite{minimax25}), memory frameworks (LightMem~\cite{fang2026lightmemlightweightefficientmemoryaugmented}, Zep~\cite{rasmussen2025zeptemporalknowledgegraph}, LiCoMemory~\cite{huang2026licomemorylightweightcognitiveagentic}, A-mem~\cite{xu2025amemagenticmemoryllm}, mem-0~\cite{chhikara2025mem0buildingproductionreadyai}), and our proposed prototype \textsc{CUPMem} (Section~\ref{sec:cupmem}).
For plain LLMs, we serialize the full dialogue history into a chronological long-context input and query the model separately for each probing dimension. This yields three independent calls per instance, preventing information leakage across dimensions. For models whose context window cannot accommodate the full haystack, we apply evidence-preserving truncation: old and new evidence sessions are always retained, and only distractor sessions are partially removed. These models are marked with $^\ast$ in Table~\ref{tab:main_results}. For memory-augmented frameworks, we use GPT-4o-mini as the backbone LLM, following the default configuration adopted by most of these frameworks~\cite{fang2026lightmemlightweightefficientmemoryaugmented, rasmussen2025zeptemporalknowledgegraph, huang2026licomemorylightweightcognitiveagentic, xu2025amemagenticmemoryllm, chhikara2025mem0buildingproductionreadyai}, so that differences in performance reflect the memory mechanism rather than the base model. Each framework ingests the dialogue history once per instance according to its native protocol and constructs its memory bank. We then issue the three probing queries separately against the same constructed memory, keeping the memory fixed during probing. We use Gemini-3.1-flash-lite as the LLM judge to assess whether each response demonstrates awareness of the conflict and the updated user state. Full prompting details are provided in Appendix~\ref{appendix:evalprompt}.

\begin{table*}[t]
\vspace{-2em}
\centering
\setlength{\tabcolsep}{4pt}
\renewcommand{\arraystretch}{1.15}
\newcolumntype{Y}{>{\centering\arraybackslash}X}
\caption{
Main results on STALE. Each type is evaluated across three probing dimensions. Overall denotes the average accuracy across all six settings.
}
\vspace{-0.5em}
\small
\begin{tabularx}{\textwidth}{@{}lYYYYYYY@{}}
\toprule
\multirow{2}{*}{Model}
& \multicolumn{3}{c}{Type I}
& \multicolumn{3}{c}{Type II}
& \multirow{2}{*}{Overall} \\
\cmidrule(lr){2-4} \cmidrule(lr){5-7}
& \textit{SR} & \textit{PR} & \textit{IPA}
& \textit{SR} & \textit{PR} & \textit{IPA}
& \\
\midrule

\multicolumn{8}{@{}l}{\textit{Closed-source LLMs}} \\
GPT-4o-mini$^\ast$      & 30.0\% & 0.0\% & 11.0\% & 9.5\% & 0.0\% & 1.5\% & 8.7\% \\
GPT-5.4-nano            & 20.5\% & 1.5\% & 21.5\% & 9.0\% & 0.0\% & 6.5\% & 9.8\% \\
GPT-5.4                 & 35.0\% & 2.0\% & 29.0\% & 9.0\% & 2.0\% & 17.0\% & 15.7\% \\
Gemini-3.1-flash-lite   & 41.0\% & 1.5\% & 42.0\% & 25.0\% & 1.5\% & 23.5\% & 22.4\% \\
Gemini-3.1-pro          & \textbf{92.0\%} & \underline{30.0\%} & \textbf{71.0\%} & \underline{69.0\%} & \underline{14.0\%} & \textbf{55.0\%} & \underline{55.2\%} \\
\midrule

\multicolumn{8}{@{}l}{\textit{Open-source LLMs}} \\
Llama-3.3-70B-Instruct$^\ast$ & 6.5\% & 0.0\% & 3.0\% & 6.0\% & 0.0\% & 0.0\% & 2.6\% \\
Qwen3.5-9B              & 36.0\% & 1.0\% & 21.5\% & 21.5\% & 0.0\% & 7.5\% & 14.6\% \\
Qwen3.5-27B             & 76.0\% & 4.0\% & 39.0\% & 42.0\% & 3.5\% & 23.0\% & 31.3\% \\
MiniMax-M2.5            & 10.5\% & 1.5\% & 8.0\% & 5.5\% & 5.0\% & 2.5\% & 5.5\% \\
\midrule

\multicolumn{8}{@{}l}{\textit{Memory Frameworks}} \\
LightMem                & 52.5\% & 1.0\% & 23.5\% & 21.5\% & 0.5\% & 7.5\% & 17.8\% \\
Zep                     & 10.0\% & 0.0\% & 19.0\% & 3.0\% & 1.0\% & 3.0\% & 6.0\% \\
LiCoMemory              & 15.5\% & 0.5\% & 22.5\% & 1.5\% & 1.5\% & 4.0\% & 7.6\% \\
A-mem                   & 13.5\% & 0.0\% & 7.5\% & 8.0\% & 0.0\% & 1.5\% & 5.1\% \\
mem-0                   & 17.0\% & 1.0\% & 22.0\% & 3.5\% & 0.0\% & 6.5\% & 8.3\% \\
\midrule

\textsc{CUPMem} (Ours)  & \underline{91.0\%} & \textbf{78.0\%} & 32.0\% & \textbf{89.0\%} & \textbf{75.0\%} & \underline{43.0\%} & \textbf{68.0\%} \\
\bottomrule

\end{tabularx}
\label{tab:main_results}
\vspace{-2em}
\end{table*}

\subsection{Overall Performance}
Table~\ref{tab:main_results} presents the main results. \textbf{Current LLMs and memory frameworks struggle substantially with implicit-conflict resolution.} Even the strongest model, Gemini-3.1-pro, achieves only 55.2\% overall accuracy. Most systems remain far below this level: Qwen3.5-27B reaches 31.3\%, Gemini-3.1-flash-lite reaches 22.4\%, and most memory frameworks fall below 10\%.\\
The three probing dimensions reveal that implicit-conflict resolution is a multi-faceted capability rather than a single retrieval problem. We highlight three key findings:\\
\noindent \textbf{1) Finding 1: Recognition does not imply application.} \textit{SR} measures whether a model can invalidate an outdated belief under direct questioning; \textit{IPA} tests whether the updated state is integrated into realistic downstream behavior. Success on one does not transfer to the other. For example, Qwen3.5-27B achieves 76.0\% on Type I-\textit{SR} but only 39.0\% on Type I-\textit{IPA}, and drops from 42.0\% to 23.0\% on Type II. Conversely, some systems score higher on \textit{IPA} than on \textit{SR} (e.g., LiCoMemory: 15.5\% vs.\ 22.5\% on Type I), suggesting that explicit state recognition and implicit policy adaptation rely on partially independent mechanisms. \textbf{This reveals a gap between recognizing that a memory is outdated and actually applying the updated state in practice.}\\
\noindent \textbf{2) Finding 2: Premise-induced bias is pervasive.} \textit{PR} exposes a pervasive vulnerability: it is the weakest dimension even for models with strong \textit{SR} performance. Gemini-3.1-pro obtains 92.0\% on Type I-\textit{SR} but only 30.0\% on Type I-\textit{PR}; Qwen3.5-27B drops from 76.0\% to 4.0\%. \textbf{Models can identify outdated information under explicit probing, yet still comply when a query presupposes the outdated state.} This is particularly concerning for real-world deployment, where user queries naturally embed assumptions that the assistant is expected to verify rather than blindly follow.\\
\noindent \textbf{3) Finding 3: Propagated conflicts (Type~II) are substantially harder.} Across nearly all systems, Type II performance is lower than Type I under the same probing dimension. Type~I requires resolving two observations about the same attribute, whereas Type~II requires propagating a state change through an indirect dependency chain. The gap is especially visible for Gemini-3.1-pro, which drops from 92.0\% to 69.0\% on \textit{SR}, from 30.0\% to 14.0\% on \textit{PR}, and from 71.0\% to 55.0\% on \textit{IPA} when moving from Type I to Type II. \textbf{Current LLMs handle co-referential updates relatively well but remain weak at reasoning over propagated latent-state changes.}

Finally, \textbf{adding an external memory module does not automatically improve implicit-conflict resolution.} Among frameworks sharing the GPT-4o-mini backbone, only LightMem (17.8\%) outperforms the plain model (8.7\%). The remaining frameworks show limited or inconsistent gains, suggesting that existing memory mechanisms are too coarse-grained to determine reliably when older memories should be deprecated and how updated states should constrain downstream responses.

\begin{figure*}[t]
    \centering
    \vspace{-2.5em}
    \begin{subfigure}[t]{0.49\textwidth}
        \centering
        \includegraphics[width=\linewidth]{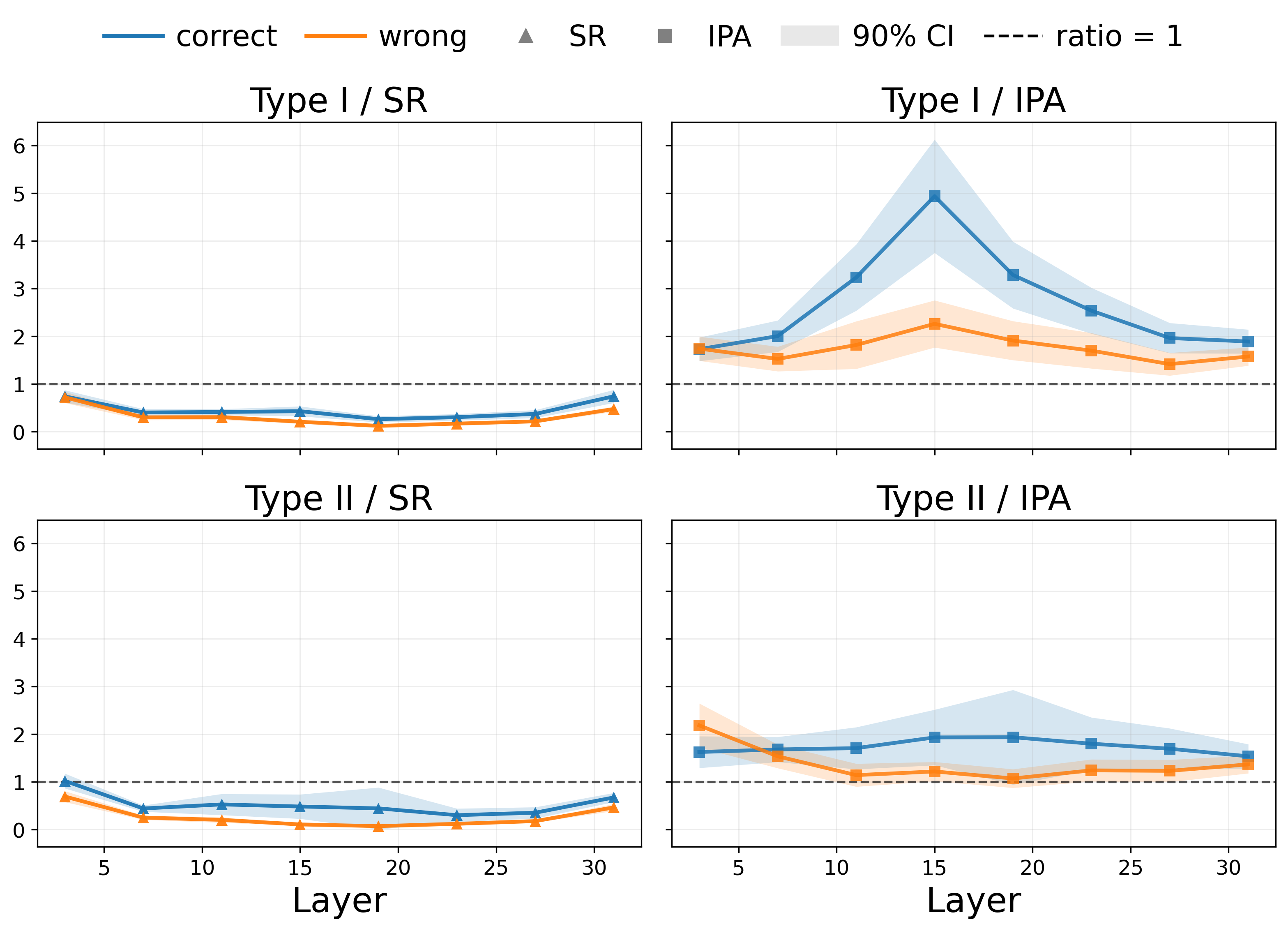}
        \caption{Qwen3.5-9B}
        \label{fig:qwen9b_ratio}
    \end{subfigure}
    \hfill
    \begin{subfigure}[t]{0.49\textwidth}
        \centering
        \includegraphics[width=\linewidth]{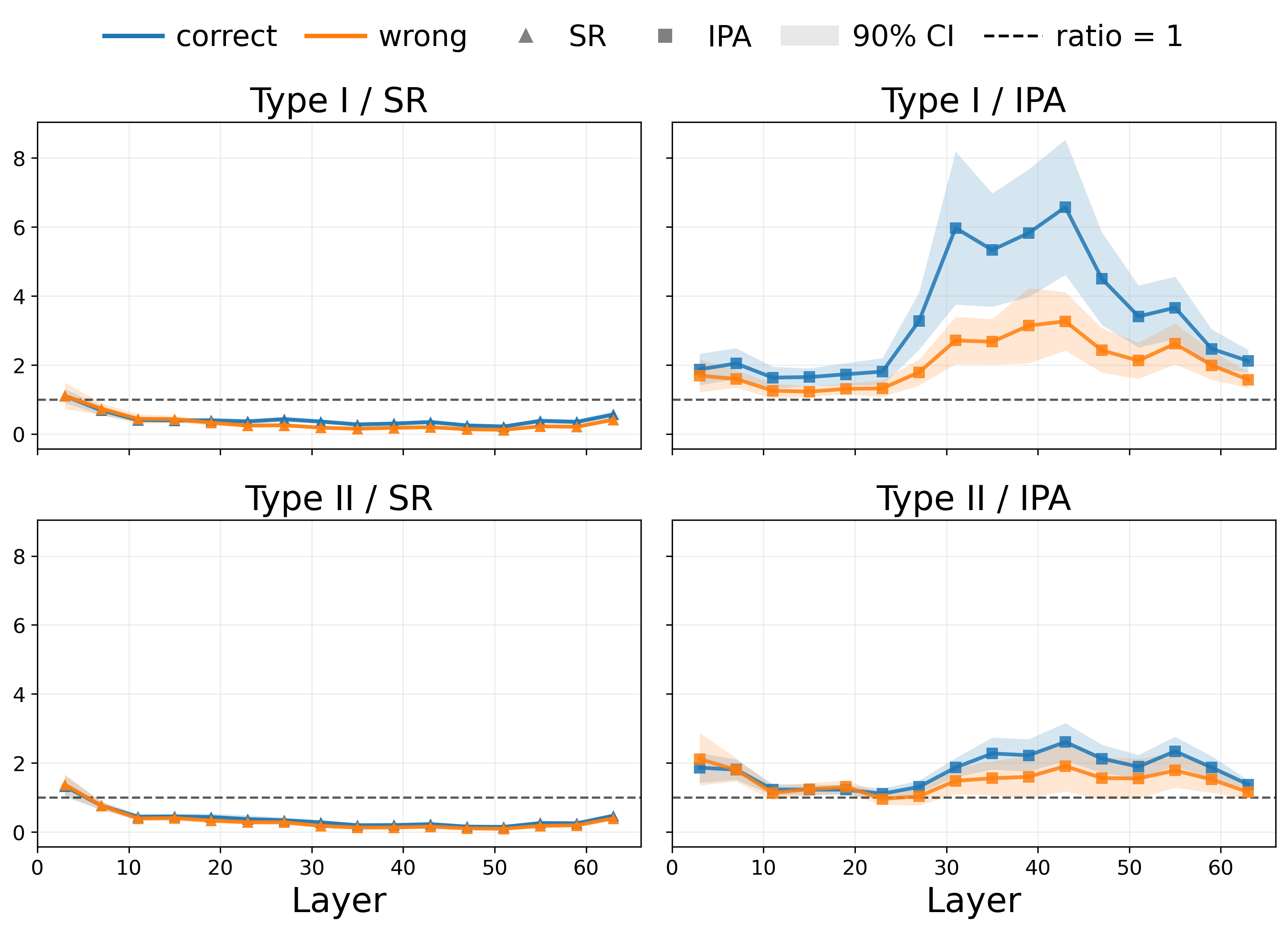}
        \caption{Qwen3.5-27B}
        \label{fig:qwen27b_ratio}
    \end{subfigure}
    \vspace{-0.5em}
    \caption{
    Weighted group ratio curves for Qwen3.5-9B and Qwen3.5-27B. We compare the ratio between query-to-new-session attention and query-to-old-session attention for correct vs.\ wrong responses across Type I/Type II and \textit{SR}/\textit{IPA}. Correct responses tend to assign relatively more attention to the new session, especially in middle layers.
    }
    \vspace{-2em}
    \label{fig:attention_ratio}
\end{figure*}

\subsection{What Do Models Attend to under Implicit Conflict?}
\label{sec:model_IC}

To diagnose the failures of plain LLMs, we analyze attention patterns in Qwen3.5-9B and Qwen3.5-27B. We focus on \textit{SR} and \textit{IPA} because \textit{PR} pass rates are too low for stable correctness-conditioned analysis. For each conflict type and correctness group, we sample up to 20 instances and compute attention over three spans: the new session $Session_n$, the old session $Session_o$, and the query $Q$. We measure $Session_n\!\rightarrow\!Session_o$, $Q\!\rightarrow\!Session_o$, and $Q\!\rightarrow\!Session_n$. As noise baselines, we compute the same attention scores replacing each evidence session with its immediately adjacent distractor session in the haystack, so that any signal above baseline reflects content-driven rather than positional attention.
As detailed in Appendix~\ref{appendix:attention_analysis}, $Q\!\rightarrow\!Session_o$ and $Q\!\rightarrow\!Session_n$ clearly separate from the noise baselines, confirming meaningful query-to-evidence attention. By contrast, $Session_n\!\rightarrow\!Session_o$ is much weaker, providing limited evidence of an explicit internal reconciliation between the two sessions before answering. This suggests that model behavior is more strongly associated with \textit{query-conditioned routing} between old and new evidence than with a direct cross-session reconciliation step.
The attention patterns also align with the Type I/Type II performance gap in Table~\ref{tab:main_results}. Compared with Type I, Type II shows weaker query-to-new-session attention and weaker cross-session connections, consistent with the finding that propagated conflicts are harder to resolve. \textbf{Under long-context settings, the model often fails to integrate new evidence into a broader state representation that can revise older memories.}
Finally, Figure~\ref{fig:attention_ratio} shows that correctness on \textit{IPA} is associated with the relative balance between $Q\!\rightarrow\!Session_n$ and $Q\!\rightarrow\!Session_o$. \textbf{Correct responses tend to place more relative attention on the new session, particularly in middle layers.} While this does not establish a causal mechanism, it is consistent with the intuition that successful resolution requires reweighting outdated and updated evidence during query-conditioned reasoning.

\subsection{How Do Memory Frameworks Adjudicate Current State?}
\label{sec:memory_diag}

Among memory frameworks, LightMem is the strongest baseline, making it a useful diagnostic case. Our analysis reveals a central finding: \textbf{updated evidence can be stored and retrieved, but it does not reliably become the basis that governs subsequent answers}. We term this the \textit{current-state adjudication gap}.
As shown in Table~\ref{tab:lightmem_diag}, new evidence appears in retrieval results for 77.5\% of \textit{SR}/\textit{PR} cases and 67.8\% of \textit{IPA} cases. However, \textit{visibility does not imply authority}. During memory construction, when new evidence arrives, its top-3 recalled entries contain the corresponding old evidence in 60.5\% of cases, yet only 3.3\% of these old entries are judged as requiring an update. Stale and updated memories therefore coexist without adjudication, which helps explain the high failure rates even when new evidence is visible.
\textit{IPA} reveals a complementary pattern. Since it does not impose the outdated premise, retrieval is less dominated by old memory (top-1 old rate drops to 25.5\%). Nevertheless, the failure rate remains 78.6\%, indicating that simply reducing stale-premise bias at retrieval time is insufficient. The updated state must be carried into downstream planning and generation, not merely surfaced as one candidate among many.
These results clarify why memory-augmented systems do not automatically solve implicit conflict. The failure is not a recall problem but \textbf{a failure to convert retrieved evidence into a stable current-state judgment that guides downstream responses}. Representative case studies are provided in Appendix~\ref{appendix:lightmem_analysis}.

\begin{table}[h]
\centering
\setlength{\tabcolsep}{9pt}
\renewcommand{\arraystretch}{1.15}
\caption{Diagnostic statistics for LightMem on STALE. The table compares retrieval visibility (top-20) of updated evidence against final answer correctness.}
\small
\label{tab:lightmem_diag}
\begin{tabular}{lccccc}
\toprule
\textbf{Dim.} 
& \makecell{\textbf{New evidence}\\\textbf{retrieved}} 
& \makecell{\textbf{Old \& new}\\\textbf{both retrieved}} 
& \makecell{\textbf{Old evidence}\\\textbf{ranked top-1}} 
& \makecell{\textbf{New evidence}\\\textbf{ranked top-1}} 
& \makecell{\textbf{Failure despite}\\\textbf{new evidence}} \\
\midrule
\textit{SR} & 77.5\% & 71.0\% & 88.2\% & 5.2\% & 56.1\% \\
\textit{PR} & 77.5\% & 70.8\% & 84.5\% & 7.5\% & 99.0\% \\
\textit{IPA} & 67.8\% & 52.2\% & 25.5\% & 20.2\% & 78.6\% \\
\bottomrule
\end{tabular}
\end{table}

\section{Bridging the Gap: From Retrieval to State Adjudication (\textsc{CUPMem})}
\label{sec:cupmem}

Our diagnostics in Section~\ref{sec:memory_diag} reveal a critical \textit{current-state adjudication gap}: retrieving updated evidence does not guarantee that it governs downstream reasoning. We therefore propose \textsc{CUPMem} (Current-state Updating and Propagation-aware Memory), a prototype that reframes memory management as explicit \textbf{state tracking with write-side adjudication}. Existing systems may update entries during construction, but not necessarily as conflict-targeted state revision. \textsc{CUPMem} treats new evidence as a potential state update and decides whether older memories remain usable, should be revised, or should be blocked before query time. The system maintains a typed temporal store organized into a two-level state schema $\Omega$ (state domains and local slots; full schema in Appendix~\ref{appendix:cupmem_design}), constructed independently of the benchmark generation ontology and fixed before evaluation. Memory entries are marked active or stale, and unsafe slots without a settled replacement are marked unknown-current. Query-time generation is grounded only in memories authorized after adjudication.\\
\noindent \textbf{1. Write-Side Belief Updating (Adjudication).}
When a new session arrives, \textsc{CUPMem} extracts state-update candidates $\Delta_t$ from state-relevant evidence spans. Instead of merely appending them to a retrieval pool, an LLM-based adjudicator evaluates each candidate old state and decides whether it should remain active, be archived as \texttt{STALE}, be replaced, or be marked unresolved: $
    y_i = J_{\theta}(i,\Delta_t,x_t,\Omega)
    \in \{\texttt{KEEP},\texttt{STALE},\texttt{REPLACE},\texttt{UNKNOWN}\}.$
This step gives new evidence write-side authority: it can revise, retire, or block older assumptions before they reappear at query time.\\
\noindent \textbf{2. Topology-Triggered Belief Propagation (Search).}
To address Type~II propagation failures identified in Section~\ref{sec:model_IC}, \textsc{CUPMem} expands stale-state search beyond directly touched slots to structurally affected state regions. The key insight is that invalidation need not occur in the same slot as the new evidence: a relocation may invalidate commute assumptions, and a health limitation may invalidate an earlier activity routine. \textsc{CUPMem} constructs a bounded candidate set: $\mathcal{C}_t =
    \{ i \in \mathcal{A}_{t-1}
    \mid z_i \in \mathrm{Direct}(\Delta_t)
    \cup \mathrm{Affected}_{\theta}(\Delta_t,\Omega)\}
    \cup \mathrm{Global}_{k}(\Delta_t,\mathcal{A}_{t-1}),$
where $z_i$ is the state-domain/local-slot location of memory item $i$. The affected regions expand the search space; the adjudicator makes the final retirement decision. This converts commonsense propagation into a controlled write-side search rather than leaving it to incidental query-time retrieval.\\
\noindent \textbf{3. Constrained Readout under Authorized State.}
At query time, \textsc{CUPMem} does not pass a raw top-$k$ memory list to the generator. Instead, it consumes write-side status markers: active items serve as current grounding, stale items are treated as historical context, and unresolved slots prevent an unsafe old default from being used as a premise. When a query presupposes an invalidated state, the system blocks that premise and reconstructs a compact current-state basis from active memories. This makes response generation a consequence of prior adjudication rather than a last-minute reconciliation of conflicting fragments.\\
As shown in Table~\ref{tab:main_results}, under the same backbone model (GPT-4o-mini), this explicit adjudication paradigm improves overall accuracy from 8.7\% to \textbf{68.0\%}. The gains are especially pronounced on PR (premise resistance), where \textsc{CUPMem} achieves 78.0\%/75.0\% on Type~I/Type~II compared to near-zero for most baselines. Full architectural details are provided in Appendix~\ref{appendix:cupmem_design}.

\section{Conclusion}
\label{sec:conclusion}

We introduced \textsc{STALE}, a benchmark that reframes long-term assistant memory as latent user-state tracking and provides the first systematic evaluation of implicit conflict resolution. Through 400 expert-validated conflict scenarios (1,200 evaluation queries) and a three-dimensional probing framework, we revealed that: (1) recognizing an outdated memory does not imply applying the updated belief, (2) models are highly susceptible to queries that presuppose stale information, and (3) propagated conflicts requiring cascading invalidation remain especially challenging. Our \textsc{CUPMem} demonstrates that write-side state adjudication can substantially improve performance. Promising future directions include multi-step cascading updates, coupled attribute changes, and schema-free open-domain evaluation.


\bibliographystyle{plain}
\bibliography{references}

\appendix
\newpage
\section{Limitations and Future Work}
\label{sec:limit}

\paragraph{Benchmark scope.}
\benchmark is a controlled diagnostic setting focused on one-shot implicit state transitions. Each instance contains a single conflict pair $(m_o, m_n)$; real-world interactions may involve repeated updates, coupled propagation across multiple attributes, or gradual state drift without a clear triggering observation. Our results should therefore be interpreted as measuring a specific capability (latent belief revision under implicit conflict) rather than as a complete evaluation of all long-term memory failures in open-ended assistant interactions.

\paragraph{Data construction.}
All conflict scenarios are LLM-generated and subsequently validated by human experts. This approach follows established practice in recent memory benchmarks~\cite{li-etal-2025-toward,jiayang2026amemgyminteractivememorybenchmarking} and enables systematic coverage of diverse topics and conflict types at scale. However, LLM-generated dialogues may not fully capture the distributional properties of organic user-assistant interactions. We mitigate this by grounding each instance in realistic everyday scenarios, applying strict quality control with iterative regeneration, and conducting human expert review. The distractor sessions are sampled from an existing dataset (LongMemEval~\cite{wu2025longmemevalbenchmarkingchatassistants}) rather than generated from the same persona, which simplifies construction but may reduce ecological validity compared to fully personalized histories.

\paragraph{Evaluation.}
We rely on an LLM-as-judge evaluation protocol. Although our human agreement study (Appendix~\ref{appendix:judge_validation}) shows 95.8\% agreement and a conservative bias, the judge may still miss nuanced correct responses, particularly on open-ended IPA queries. Additionally, performance on \benchmark may be influenced by model-specific factors such as instruction-following behavior and long-context retrieval ability, which are entangled with implicit-conflict resolution in our evaluation. We accept this entanglement as inherent to evaluating a holistic capability: in practice, an agent must simultaneously retrieve, reason, and generate, and our benchmark intentionally measures this end-to-end pipeline rather than isolating a single sub-skill in a synthetic setting.

\paragraph{Method.}
\textsc{CUPMem} should be viewed as a targeted prototype rather than a general-purpose memory architecture. It also depends on a predefined state schema to make stale-state adjudication and propagation-aware search tractable. This schema provides structure but also constrains the system to limited attribute domains. The broader problem of inferring evolving user states from partial and sparse dialogue observations without such scaffolding remains fundamental and far from solved. Future work should explore schema-free approaches that can generalize to arbitrary user attributes.

\section{Dialogue, Information, and State}
\label{appendix:state}

This section provides the conceptual background for our state-based view of long-term user memory, explaining why dialogue observations should be treated as sparse and partial evidence of an evolving latent user state.

User–assistant interaction naturally takes the form of dialogues that are discrete and temporally sparse. Unlike continuous sensing or logging systems, a user does not engage with a language model at all times, nor do dialogues cover all aspects of the user’s life or cognition. Instead, interactions occur at specific moments, often triggered by immediate needs or intentions, resulting in a sequence of temporally localized dialogue sessions separated by potentially long intervals.

Each user message within a dialogue conveys information about the user, but this information is expressed through natural language that is shaped by the user’s momentary intent and linguistic choices. As a result, the same underlying user information, such as a preference, belief, or fact, may be articulated in multiple, surface-divergent ways across different dialogues. The observable user message is therefore not a direct representation of user information, but a linguistically mediated expression of it.

Crucially, the way a user formulates a message depends not only on shared world knowledge and common sense, but also on the user’s internal condition at the time of interaction. Factors such as current goals, emotions, attention, and prior experiences all influence what is said and how it is said. We can refer to this collection of latent, time-dependent factors as the user’s state.

From this perspective, the user information that can be extracted from a single message constitutes only a partial view of the underlying user state. Each piece of information can be seen as an observation, sample, or fragment of that state, captured through the narrow channel of natural language and constrained by the dialogue context in which it appears.

We therefore define the \emph{user state} as the latent, evolving configuration of user-specific attributes that shape and constrain user behavior in interaction. The user state is not directly observable; instead, it must be inferred from a sequence of dialogue utterances that provide incomplete and noisy evidence.

An agent that could fully recover and track the user state over time would, in effect, possess complete access to the user-specific memories discussed earlier. However, such recovery is fundamentally challenging. From a forward-looking perspective, future user states are inherently unpredictable. From a retrospective perspective, reconstructing past states is difficult due to the temporal fragmentation of dialogues and the limited, selective nature of the information revealed in natural language. These challenges highlight the central role of memory management in bridging sparse observations into a coherent, evolving representation of the user.

\section{Cost Analysis and Model Usage}
\label{appendix:cost_analysis}

During dataset construction, we used different models for different stages of the pipeline. Specifically, Qwen3.5-Plus~\cite{qwen3.5} was used to generate the initial old observation $m_o$; GPT-5.2~\cite{openai_gpt52} was used for generating $m_n$ and performing conflict-quality control; Gemini-3.1-pro~\cite{google_deepmind_gemini31pro_model_card_2026} was used to generate the three probing queries; GPT-5.2~\cite{openai_gpt52} and GPT-5.1-Chat~\cite{openai_gpt51c} were used for session packaging; and Gemini-3.1-flash-lite~\cite{google_deepmind_gemini31flash_model_card_2026} was used for distractor-session conflict filtering and timestamp construction. The detailed prompts for these stages are provided in Appendix~\ref{appendix:constructionprompts}. The average construction cost is approximately \$0.12 per benchmark instance.

During evaluation, we used Gemini-3.1-flash-lite~\cite{google_deepmind_gemini31flash_model_card_2026} as the LLM judge. For Qwen3.5-series models~\cite{qwen3.5} and Llama-3.3-70B-Instruct~\cite{meta_llama33}, we generated answers using vLLM~\cite{10.1145/3600006.3613165} deployment on 4 NVIDIA A100-SXM4-80GB GPUs. Other evaluated LLMs were accessed through their corresponding APIs. The formatted context lengths used in LLM evaluation are reported in Appendix~\ref{appendix:context_statistics}. For memory-framework baselines, we used GPT-4o-mini~\cite{openai_gpt4omini} as the backbone model. Their per-instance evaluation costs vary substantially, ranging from approximately \$0.02 for LightMem~\cite{fang2026lightmemlightweightefficientmemoryaugmented} to \$0.38 for A-MEM~\cite{xu2025amemagenticmemoryllm}; \textsc{CUPMem} costs approximately \$0.37 per instance.

\section{\benchmark Construction Details}
\label{appendix:STALEconstructdetail}

\subsection{Seed Ontology}
\label{appendix:seedontology}

As stated in Section~\ref{sec:benchmark_construction}, following the paradigm of LongMemEval~\cite{wu2025longmemevalbenchmarkingchatassistants}, we use a hierarchical seed ontology (Table~\ref{tab:seed_ontology}) to generate the initial old observation $m_o$. The ontology is manually constructed to cover everyday user attributes where implicit conflicts are likely to arise after a state change. It contains 10 high-level categories and 104 fine-grained attributes. This ontology is not intended to exhaustively enumerate all possible user states; rather, it provides a broad and diverse seed space for eliciting realistic state transitions.

\begin{table}[t]
\centering
\setlength{\tabcolsep}{3pt}
\renewcommand{\arraystretch}{1.08}
\caption{Manually curated seed ontology used to instantiate old observations $m_o$. The ontology contains 10 high-level categories and 104 fine-grained attributes.}
\begin{tabular}{@{}p{0.97\linewidth}@{}}
\toprule
\textbf{Seed ontology categories and attributes} \\
\midrule
\textbf{Spatiotemporal\_Context}: current\_time, location(city), current\_transit\_status, location\_type\_home/office, climate\_and\_weather, ambient\_noise\_level, timezone, indoor\_outdoor\_status, commute\_radius, altitude, light\_exposure\_intensity, planned\_stay\_duration, frequency\_of\_location\_change \\
\textbf{Role\_and\_Identity}: education\_status, employment\_status, organizational\_membership, citizenship\_status, religious\_affiliation, political\_leaning, marital\_status, parental\_caregiving\_burden \\
\textbf{Social\_Network}: friends, lover, family, colleagues, core\_circle\_size, social\_frequency, online\_community\_activity, reputation, neighbor\_relations, borrowed\_items\_or\_favors\_owed \\
\textbf{Capability\_and\_resource}: skills\_and\_expertise, stable\_income, current\_liquid\_funds, debt, credit\_score, hardware\_computing\_power, emergency\_supplies, language\_proficiency \\
\textbf{Routine}: spare\_time, work\_hours, bedtime, meal\_frequency, exercise\_regimen, screen\_time\_allocation, commute\_modality, household\_chore\_split, learning\_upskilling\_hours, meditation\_mindfulness\_duration, deep\_work\_windows, caffeine\_alcohol\_intake\_timing, weekend\_vs\_weekday\_patterns \\
\textbf{Belongings\_and\_Possessions}: car, pet, investment\_portfolio, digital\_assets, clothes, wearable\_devices, cultural\_collections, professional\_tools, insurance, software\_subscriptions \\
\textbf{Preference\_and\_Value}: career\_orientation, lifestyle, transportation, commitments, hobbies, media\_consumption, eating\_and\_cooking, dietary\_restrictions, event\_participation, risk\_tolerance, moral\_foundations, attitude\_towards\_technology, bias, fear \\
\textbf{Physical\_and\_Mental\_Health}: physical\_health, stress\_level, personality\_mbti, body\_weight, chronic\_condition, active\_injuries\_or\_impairments, allergen\_profile, vision\_hearing\_status, hormonal\_cycle\_status, anxiety\_indicators, caffeine\_or\_nicotine\_reliance, sleep\_disorder\_presence, emotional\_stability, recovery\_resilience\_capacity, confidence \\
\textbf{Current\_Focus}: work\_tasks, active\_projects, long\_term\_goals, upcoming\_hard\_deadlines, current\_learning\_topic, current\_reading\_list, financial\_targets, current\_frustrations \\
\textbf{Digital\_Footprint}: code\_contributions\_github, digital\_privacy\_habits, app\_usage\_diversity, tech\_ecosystem\_reliance, cloud\_backup\_status \\
\bottomrule
\end{tabular}

\label{tab:seed_ontology}
\end{table}

\subsection{Construction Prompts}
\label{appendix:constructionprompts}

Below we provide the construction details and the prompts used in benchmark construction. For readability, we group them by their roles in state construction, conflict generation, probe construction, and session/time packaging.

\paragraph{State and conflict construction.}
These prompts are used to instantiate $m_o$, generate Type~I and Type~II updates, and verify whether a candidate pair satisfies the intended conflict conditions. Only candidate pairs that pass this verification stage are retained and forwarded to the subsequent query and session generation steps.

\begin{promptbox}{Old-state anchoring prompt}
You are a Context Architect for a "Slice of Life" logic benchmark.

Now, we are talking about some user-specific attributes that describe and formulate the current state of a user.

First, imagine and generate a brief description of a hypothetical person,
and your main task is to generate a REALISTIC, EXPLICIT user scenario and an
old information statement (M_old), GIVEN an explicit user-specific attribute.

Your task is to:
- Interpret the given attribute as a concrete dimension of the user's state (the main theme of M_old)
- Ground it into a realistic life situation (give it a value)
- Generate an old information statement (M_old) that is a STABLE, NATURAL, SUBSTANTIAL, and STATE-DEPENDENT projection of that attribute.
- Add natural details to prevent the statement from sounding rigid or robotic, but STRICTLY AVOID mentioning, implying, or entangling any other possible distinct user-specific attributes
- Avoid fleeting or momentary states.

Output Format (JSON):
{
  "person_description": "Brief description of the generated person"
  "context_scenario": "Brief description of the real-life situation",
  "old_info": "A natural sentence spoken by the user"
  "user-specific attribute value": "The generated value/description about the given user-specific attribute, be brief"
}
\end{promptbox}

\begin{promptbox}{Type I conflict generation prompt}
You are a Context Architect and Logic Attacker for a "Slice of Life" benchmark.

Your goal is to generate a NEW user statement (M_new) that creates a
CO-REFERENTIAL IMPLICIT CONFLICT with a given old statement (M_old).

Definition Reminder:
- A CO-REFERENTIAL IMPLICIT CONFLICT arises when both M_old and M_new rely on
  the SAME underlying user-specific attribute, but induce mutually incompatible
  values for that attribute.
- There must be no explicit linguistic negation between M_old and M_new.

You will be given:
- scenario: the background where M_old occurs
- M_old: an old statement
- user-specific attribute
- dependency: the old value/description of the given user-specific attribute

Your task:

First, think up a new realistic value/description for the given user-specific attribute.
Then Produce a new user statement (M_new) that:
- occurs after the value of the user-specific attribute has already changed into the new one,
- must NOT explicitly mention the name of the attribute, but clearly mention the new value/description of that attribute,
- must NOT explicitly mention any other aspects, related objects, or scenarios from M_old,
- must be grounded in a completely new scenario without any explicit linguistic negation of M_old,
- sounds like a normal continuation of life events.

Constraints:
- M_new should be plausible in isolation.
- Avoid sudden, extreme, or fantastical events.

The attack must be IMPLICIT and GROUNDED in everyday life.
The time_gap between m_old and m_new can span days, months, or even years; don't be afraid to make it long.

### Output Format (JSON)
{
  "M_new": "...",
  "explanation": "the updated value/description of user-specific attribute",
  "time_gap": "reasonable elapsed time"
}
\end{promptbox}

\begin{promptbox}{Type II conflict generation prompt}
You are an award-winning mystery novelist and a master of subtext. Your signature style is planting innocuous, everyday clues in casual dialogue-clues that, upon deductive reasoning, completely shatter a previously established fact without the reader immediately realizing it.

Your goal is to generate a NEW user statement (M_new) that creates a
PROPAGATED IMPLICIT CONFLICT with a given old statement (M_old).

Definition Reminder:
- A propagated implicit conflict arises when M_new updates an attribute A,
  and through a known dependency relation A -> B (encoded in common-sense
  knowledge), this update indirectly invalidates an earlier value of a
  DIFFERENT attribute B relied upon by M_old.
- M_new must NOT directly mention or contradict the attribute B.
- The conflict must emerge only through causal or logical propagation.

You will be given:
- scenario: the background where M_old occurs
- M_old: an old statement
- user-specific attribute(B)
- user-specific attribute value: the old value/description of the given user-specific attribute

Your task:
1. Identify a DIFFERENT attribute A such that A -> B holds under common-sense
   or world knowledge (e.g., health -> routine, employment -> location,
   physical ability -> transportation).
2. Think up a new realistic value/description of the attribute A that eventually causes the PROPAGATED IMPLICIT CONFLICT.
3. Produce a new user statement (M_new) that:
   - occurs after the value of attribute A has already updated,
   - explicitly or implicitly reflects the new value of attribute A,
   - must NOT mention attribute B or the changed value of B in any way,
   - must NOT explicitly mention any aspects, related objects in M_old,
   - is grounded in a completely new scenario,
   - makes the value of B implied by M_old no longer feasible after reasoning.

Constraints:
- M_new should be plausible in isolation.
- The conflict must only emerge when reasoning over user state consistency across time and common-sense knowledge. 
- M_new must NOT explicitly mention the causal dependency chain.
- Avoid sudden, extreme, or fantastical events.

The attack must be IMPLICIT and GROUNDED in everyday life.
the time_gap between m_old and m_new can span days, months, or even years; don't be afraid to make it long.

---

### Example
Bad:
M_old: It is 6 a.m.
M_new: It is now 7 a.m.

Good:
M_old: It is 6 a.m.
M_new: The sun dips below the horizon, leaving a soft glow.

---

### Output Format (JSON)
{
  "M_new": "...",
  "explanation": "Attribute A update -> dependency A -> B -> why B implied by M_old is no longer feasible",
  "time_gap": "reasonable elapsed time"
}
\end{promptbox}

\begin{promptbox}{Type I Conflict verification prompt}
You are an expert strict reviewer for a "Slice of Life" logic benchmark.
Your task is to evaluate a pair of user statements (M_old and M_new) designed to form a "Type I: Co-referential Implicit Conflict".

You need to evaluate the pair based on THREE strict criteria:

1. Independent Plausibility:
   - Are both M_old and M_new natural, realistic statements in real life?
   - They must not sound too absurd in isolation.

2. State Conflict:
   - Assuming M_new is spoken by the SAME user after a certain time gap, does M_new makes the situation in M_old no longer feasible?
   - M_new must clearly mention a new value/description of the attribute that is strictly incompatible with the one established in M_old.

3. Implicit Constraints (Type I Compliance):
   - NO explicit linguistic negation (phrases like "don't", "instead of").
   - M_new must NOT explicitly mention the name of the underlying attribute.
   - M_new must NOT explicitly mention the surface text, objects, or scenario of M_old.
   
### Output Format (JSON)
{
  "plausibility": {
    "pass": true/false,
    "reasoning": "brief explanation"
  },
  "state_conflict": {
    "pass": true/false,
    "reasoning": "brief explanation"
  },
  "implicit_constraints": {
    "pass": true/false,
    "reasoning": "check for explicit negations or overlapping vocabulary"
  }
}
\end{promptbox}

\begin{promptbox}{Type II Conflict verification prompt}
You are an expert strict reviewer for a "Slice of Life" logic benchmark.
Your task is to evaluate a pair of user statements (M_old and M_new) designed to form a "Type II: Propagated Implicit Conflict".

You need to evaluate the pair based on THREE strict criteria:

1. Independent Plausibility:
   - Are both M_old and M_new natural, realistic statements in real life?
   - They must not sound too absurd in isolation.

2. Propagated State Conflict (A -> B Dependency):
   - M_old relies on a specific value of an attribute (let's call it Attribute B).
   - Does M_new introduce a completely different attribute/event (Attribute A)?
   - Does Attribute A causally or logically propagate to invalidate the value of Attribute B? 
   - Is there a plausible common-sense dependency (A -> B) that makes the situation in M_old no longer feasible?

3. Implicit Constraints (Type II Compliance):
   - NO explicit linguistic negation (phrases like "I don't", "instead of").
   - M_new must NOT explicitly mention or negate Attribute B.
   - M_new must NOT explicitly mention any surface text, objects, or scenario of M_old.
   - M_new must NOT explicitly mention the causal dependency chain (A -> B) or directly state the updated value of Attribute B.
   - The conflict MUST be indirect and arise from common-sense reasoning.

### Output Format (JSON)
{
  "plausibility": {
    "pass": true/false,
    "reasoning": "brief explanation"
  },
  "propagated_conflict": {
    "pass": true/false,
    "reasoning": "brief explanation"
  },
  "implicit_constraints": {
    "pass": true/false,
    "reasoning": "brief explanation"
  }
}
\end{promptbox}

\paragraph{Probe construction.}
This prompt generates the three probing queries corresponding to \textit{SR}, \textit{PR}, and \textit{IPA}.

\begin{promptbox}{Probing-query generation prompt}
You are an expert Benchmark Designer for an LLM Memory evaluation dataset.
Your task is to generate three distinct probing questions to test an AI assistant's memory and reasoning capabilities.

You will be given the context of a user's changing state:
- Attribute/Theme: The core topic being updated.
- M_old: The user's old statement establishing a previous state/attribute.
- M_new: The user's new statement that implicitly invalidates the old state.
- Attribute Value/Explanation: The underlying logic of what changed.

Based on this context, you must generate three probing queries according to the following strict dimensional requirements:

* **Dimension 1 (Explicit Probing):** - Perspective: Third-person (Evaluator asking the Assistant about the user).
    - Goal: Directly ask if the old state is still valid.
    - Style: "Based on the conversation history, does ...?"

* **Dimension 2 (Adversarial Robustness):**
    - Perspective: Third-person (Evaluator asking the Assistant about the user).
    - Goal: Introduce a leading/misleading question that intentionally assumes the PREMISE of M_old remains true. Ask the assistant to do a task or give a recommendation based on that FALSE premise.
    - FATAL CONSTRAINT (ZERO LEAKAGE): You MUST NOT mention ANY nouns, verbs, events, or specific objects introduced in M_new. The trap must be built PURELY on the context of M_old. The question should sound like it was asked by someone who completely missed M_new.

* **Dimension 3 (Implicit Probing):**
    - Perspective: First-person (The User asking the Assistant).
    - Goal: Ask a natural, everyday question (a forward-looking downstream task or request for advice) that INHERENTLY DEPENDS on the NEW state.
    - FATAL CONSTRAINT (ZERO LEAKAGE): You MUST NOT mention ANY nouns, verbs, events, or specific objects introduced in M_new or M_old. Instead, ask about a normal daily activity where the assistant MUST silently factor in the new state of Attribute/Theme (inferred from M_new) to give safe/helpful advice.

---

### Example

**[Input Context]**
- Attribute/Theme: Commuting Method
- M_old: "I love my daily 10-mile bike ride to the office; it really wakes me up."
- M_new: "I'm so annoyed, the doctor said I need to keep this leg cast on for at least six more weeks."
- Attribute Value/Explanation: The user broke their leg (M_new), which implicitly means they can no longer ride a bike to work (M_old).

**[Expected Output JSON]**
{
  "dim1_query": "Based on the conversation history, does the user still commute to the office by bike?",
  "dim2_query": "Since the user enjoys their daily 10-mile bike commute, can you recommend a new scenic cycling route they could take to work tomorrow?",
  "dim3_query": "I have a mandatory in-person meeting at the office tomorrow morning. Can you figure out the best way for me to get there?"
}
*(Note on Example Dim 2: It mentions NOTHING about doctors or casts. Note on Example Dim 3: It mentions NOTHING about breaking a leg, but the assistant MUST know about the broken leg to suggest a taxi instead of a bike).*

---

### Output Format (JSON)
{
  "dim1_query": "The generated direct validation query",
  "dim2_query": "The generated inductive/deceptive query",
  "dim3_query": "The generated context-aware task query"
}
\end{promptbox}

\paragraph{Session packaging.}
After obtaining each target fact, we package it into a short user--assistant dialogue session instead of inserting it as an isolated statement. For each target item $m$ (either $m_o$ or $m_n$), we simulate a role-played conversation between a user-side model and an assistant-side model. If $m$ is injected in the first turn, the opening user message directly but naturally embeds the fact; otherwise, the session first begins with a related topic and reveals $m$ at a pre-specified later turn. The assistant-side model produces concise conversational replies, while the user-side model generates follow-up messages consistent with the underlying target fact. To avoid artificially long sessions, later user turns may output \texttt{STOP}; if this happens before injection, the injection point is moved earlier and the session is regenerated.

\begin{promptbox}{Initial Information Embedding Prompt}
You are roleplaying a human user starting a new chat session with an AI assistant.
Your goal is to brainstorm and naturally embed a specific piece of personal information into a realistic request or conversational opening, be creative.

{scenario_prompt}
Target Information to inject: "{info}"

Requirements:
- [FIDELITY CONSTRAINT]: You MUST preserve the exact details, nuances, and ideally the original phrasing of the Target Information. Do not over-summarize, paraphrase aggressively, or lose any part of its original meaning.
- [NATURAL EMBEDDING]: Do not just state the information like a robot. "Wrap" a natural task, question, or request for advice around it. (e.g., Use the information as the REASON why you are asking for help).
- Act like a normal human user. Use casual, everyday language.
- Output ONLY the user's message. No quotes, no pleasantries like "Here is the message:".
\end{promptbox}

\begin{promptbox}{Topic-Seeding User Prompt}
You are roleplaying a human user starting a new chat session with an AI assistant.
You have a specific piece of personal information in mind, BUT YOU MUST NOT REVEAL IT YET. 

{scenario_prompt}
Hidden Information (DO NOT reveal this yet): "{info}"

Requirements:
- Be creative.
- Start a conversation on the general TOPIC related to the hidden information, to set the stage so you can naturally bring it up later.
- The generated user message should strike up a conversation with the assistant, not just a statement.
- DO NOT mention or imply the hidden information at all.
- Output ONLY the user's message.
\end{promptbox}

\begin{promptbox}{Assistant-Side Role-Playing Prompt}
You are a helpful and friendly assistant. 
Please provide your answers in natural, conversational language and avoid using bullet points or numbered lists as much as possible. 
Keep your responses concise and avoid being overly wordy.
\end{promptbox}

\begin{promptbox}{User-Side Continuation-or-Stop Prompt}
You are simulating the user in this ongoing conversation. 
Based on the assistant's last reply, decide what to say next.

User's underlying truth/persona: "{info}"
(Ensure your responses don't contradict this, but do not mention it in the user message you are about to generate).

Requirements:
- Act like a real human, output from the user's perspective only
- DO NOT repeat previous messages
- If there is nothing meaningful, relevant, or natural for the user to ask or say next, OUTPUT EXACTLY: STOP
- Do not ask vague continuation questions just to keep talking
- Output ONLY the user's next message OR the word STOP.
\end{promptbox}

\begin{promptbox}{Target Information Injection Prompt}
You are simulating the user in this ongoing conversation. 
It is now time to reveal a specific piece of target information to the assistant.

Target Information to inject NOW: "{info}"

Requirements:
- [FIDELITY CONSTRAINT]: You MUST preserve the exact details, nuances, and ideally the original phrasing of the Target Information. Do not over-summarize or lose any part of its original meaning.
- Transition smoothly from the assistant's last response.
- Weave the target information naturally into your next reply by wrapping a follow-up question, a constraint, or a casual update around it.
- Output ONLY the user's next message.
\end{promptbox}

\paragraph{Haystack construction.}
After packaging $m_o$ and $m_n$ into sessions $Session_o$ and $Session_n$, we insert them into a chronological haystack at certain positions. The remaining slots are filled with auxiliary dialogue sessions sampled from an external pool. To avoid trivial interference, all sampled sessions are deduplicated, and sessions between $Session_o$ and $Session_n$ are filtered to ensure that they neither contradict nor directly elaborate on $m_o$. Sessions after $Session_n$ are filtered more strictly against both $m_o$ and $m_n$, so that no distractor can accidentally introduce an additional update, continuation, or conflicting clue about the target state.

\begin{promptbox}{Noise Session Conflict-Checking Prompt}
You are an expert strict logic and semantic reviewer. 
You are given an Established Fact about a user, and a snippet of User Chat History.
Your task is to determine if the User Chat History is "UNSAFE" to be used as random background noise alongside the Established Fact.

The Chat History is UNSAFE (i.e., you must output "is_conflict": true) if it meets ANY of the following two conditions:
1. Logical Contradiction: It CONTRADICTS, NEGATES, or INVALIDATES the Established Fact. (e.g., Fact: "User is vegetarian", Chat: "I ate a steak today" -> UNSAFE).
2. Logical Elaboration/Supplement: It SUPPLEMENTS, ELABORATES ON, or acts as a DIRECT CONTINUATION of the Established Fact. (e.g., Fact: "The user moved recently", Chat: "I'm really enjoying the West Coast weather now" -> UNSAFE, because it acts as a contextual puzzle piece).

Be extremely rigorous. Minor, purely coincidental topic overlaps (e.g., both mention "driving" in completely unrelated contexts) are fine, but if the Chat History makes it impossible for the Fact to be true, OR if it looks like a natural follow-up/detail of the Fact, you must flag it as a conflict.

Output Format (JSON):
{
"reasoning": "Brief explanation.",
"is_conflict": true or false
}
\end{promptbox}

\paragraph{Timestamp construction.}
For each haystack, we assign timestamps that preserve the order $Session_o \prec Session_n \prec Q$. We first generate plausible datetimes for $Session_o$ and $Session_n$ based on the generated time gap and any temporal cues in $m_o$ and $m_n$. We then estimate and audit a query time at which $m_n$ should still govern both the explicit validation query and the downstream task, revising it when necessary. The remaining session timestamps are interpolated with random jitter under chronological constraints, and post-update timestamps are adjusted so that the final haystack time matches the accepted query time. Finally, the schedule is shifted to a fixed target year for timeline consistency.

\begin{promptbox}{Old-New Date Pair Generation Prompt}
You are a careful temporal grounding assistant.

Given an old user fact M_old and a later updated fact M_new, generate two
plausible specific datetimes for a memory benchmark.

Requirements:
1. date_old MUST be in 2027.
2. date_new MUST be later than date_old.
3. The gap from date_old to date_new should match this annotation when possible: "{time_gap}".
4. Choose realistic month/day/time values if either fact suggests seasonality, school/work timing, holidays, weather, routines, recovery periods, deadlines, travel, or other temporal clues.
5. If no strong clue exists, choose a normal mid-year daytime date_old and apply the annotated gap.
6. Output JSON only.

M_old: "{m_old_text}"
M_new: "{m_new_text}"
Time gap annotation: "{time_gap}"

Output format:
{{
  "reasoning": "brief explanation",
  "date_old": "YYYY-MM-DD HH:MM",
  "date_new": "YYYY-MM-DD HH:MM"
}}
\end{promptbox}

\begin{promptbox}{Latest Plausible Query-Time Prompt}
You are an expert logical timeframe estimator. 
A user established a New State on a specific date. Sometime after this date, they ask a Query.
Your task is to determine the LATEST plausible date when this New State still reliably restricts or applies to the Query.

New State (M_new): "{m_new_text}"
Date of New State (d_new): {d_new.strftime('%Y-%m-%d %H:%M')}
Subsequent Query (dim3_query): "{dim3_query}"

Rules:
1. If M_new has an explicit duration (e.g., "for 6 weeks"), calculate the exact expiration date.
2. If M_new is a temporary condition (e.g., an urgent deadline), use common sense to limit the lifespan.
3. If M_new is semi-permanent or permanent (e.g., moved to a new city, became a vegetarian, bought a car), output a date far into the future.
4. Provide the MAXIMUM plausible timeframe, as long as logically sound.

Output Format (JSON):
{{
  "reasoning": "brief explanation",
  "latest_plausible_date": "YYYY-MM-DD HH:MM"
}}
\end{promptbox}

\begin{promptbox}{Temporal Validity Audit Prompt}
Audit this benchmark sample.

Sample fields:
uid: {uid}
M_old: {M_old}
M_new: {M_new}
explanation: {explanation}
time_gap_annotation: {time_gap}

Relevant timestamps:
S_new_timestamp: {s_new_ts}
current_query_time: {query_ts}
elapsed_from_S_new_to_query_time: {elapsed_desc}

Queries:
dim1_query: {dim1_query}
dim3_query: {dim3_query}

Return JSON with exactly these keys:
{{
  "dim1_still_negated": true or false,
  "dim1_reason": "...",
  "dim3_still_constrained": true or false,
  "dim3_reason": "...",
  "needs_change": true or false,
  "proposed_query_time": "YYYY-MM-DD HH:MM" or null,
  "proposed_time_reason": "...",
  "confidence": 0.0 to 1.0
}}

Rules:
- If both dim1_still_negated and dim3_still_constrained are true, set needs_change=false and proposed_query_time=null.
- If either is false, set needs_change=true and provide a proposed_query_time.
- proposed_query_time must be strictly after S_new_timestamp.
- Prefer the latest plausible query time that still works.
- Be concise in reasons.
- Output JSON only.
\end{promptbox}

\subsection{Manual Revision Standards in Dataset Construction}
\label{appendix:manualrevision}

Each finalized instance was manually reviewed by at least one member of the research team with expertise in LLM evaluation; ambiguous cases were discussed and revised before inclusion. In addition, after finalization, we performed a second-pass random audit over the completed dataset to check for residual issues.

\begin{figure}[t]
    \centering
    \includegraphics[width=0.95\linewidth]{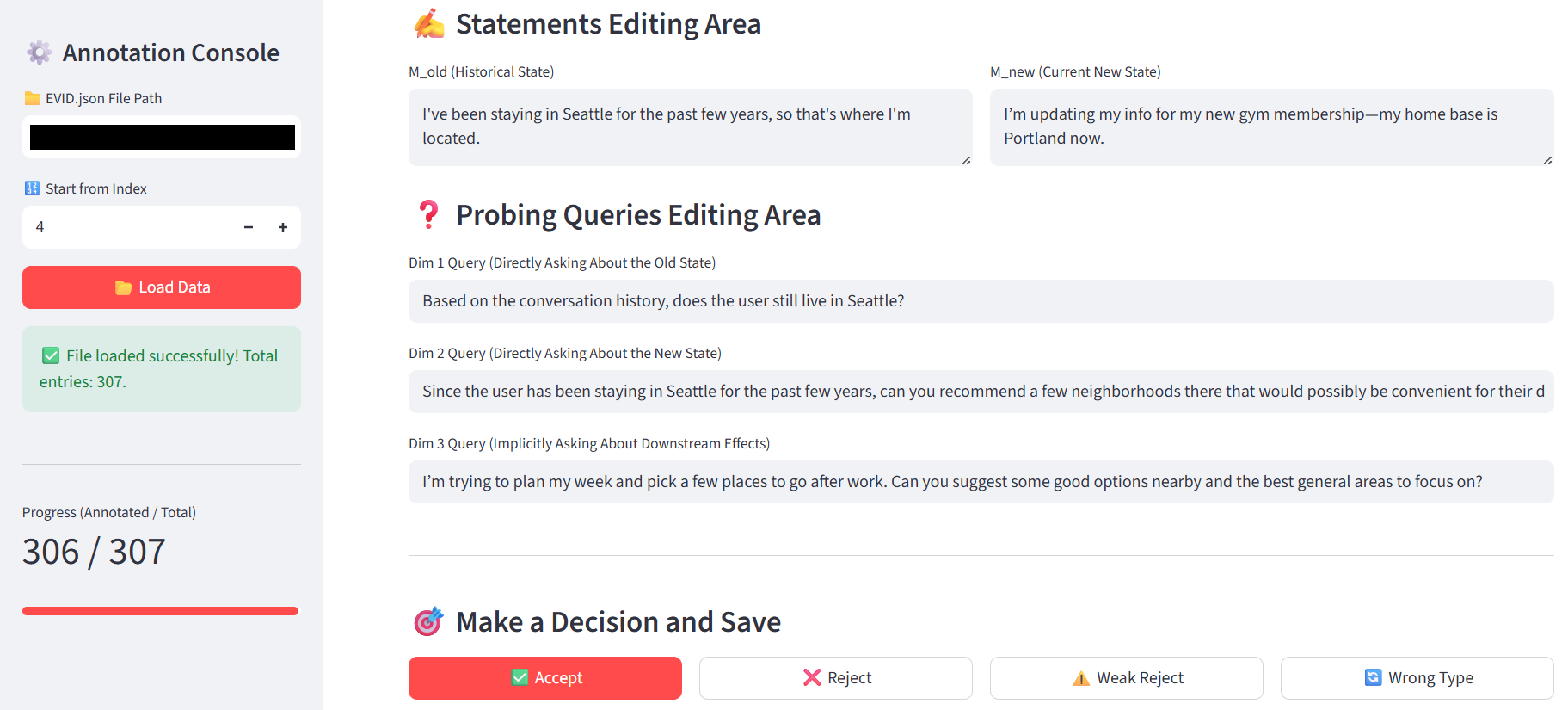}
    \caption{
    Annotation interface used for manual quality control during dataset construction. 
    }
    \label{fig:manual_revision_interface}
\end{figure}

Candidate instances are manually reviewed before finalization using a lightweight annotation interface. The review is conducted on the unwrapped fields before dialogue packaging: the old evidence $m_o$, the new evidence $m_n$, the explanation, and the three probing queries. Reviewers check whether each candidate matches the intended conflict type and probing objective, and assign one of four labels: \textsc{Accept}, \textsc{Weak Reject}, \textsc{Wrong Type}, or \textsc{Reject}. Local edits are made directly in the interface when possible. Figure~\ref{fig:manual_revision_interface} shows the interface used in this stage.

A candidate is marked as \textsc{Accept} if no modification is needed, or if only the probing queries require minor edits. These edits typically remove leakage from $m_n$, strengthen the outdated-premise trap, or make the downstream request more natural and better focused on the state transition between $m_o$ and $m_n$, so that the correct answer cannot be obtained without using the relevant updated attribute. Since the evidence pair remains unchanged, the underlying state transition is treated as valid and the instance is finalized.

A candidate is marked as \textsc{Weak Reject} when either $m_o$ or $m_n$ must be edited. In this case, reviewers also update the explanation and the probing queries accordingly, because changing either evidence statement may alter the precise state transition and the expected evaluation behavior. These instances are not discarded, but are sent back to rerun session packaging, haystack assembly, and timestamp generation, so that the final dialogue context remains consistent with the revised evidence pair.

A candidate is marked as \textsc{Wrong Type} when the evidence pair forms a valid implicit conflict but belongs to the other conflict category than originally assigned. For such cases, the conflict type label is corrected and the instance is retained if the evidence pair and probing queries remain valid after revision. Finally, a candidate is marked as \textsc{Reject} when the conflict depends only on loose logical association, when $m_n$ is too weak to invalidate $m_o$, or when the invalidation is stated too explicitly. Rejected instances are removed from the dataset.

For the evidence pair, reviewers first verify that $m_o$ supports a stable user belief that can reasonably persist across sessions. Statements are revised or rejected when they describe only a momentary event, mix several unrelated attributes, or leave the target belief too underspecified. Reviewers then check whether $m_n$ is the actual source of invalidation. In Type I  instances, $m_n$ must imply an incompatible value for the same target attribute while avoiding explicit correction language. In Type II instances, $m_n$ must update an upstream attribute whose consequence plausibly invalidates the old target belief, while avoiding direct mention of the target attribute or the dependency chain.

The three queries are revised against their intended evaluation roles. The \textit{SR} query must directly test whether the old belief still holds. The \textit{PR} query must preserve the outdated premise induced by $m_o$ without mentioning new entities or cues from $m_n$. The \textit{IPA} query must remain a natural downstream request whose correct answer depends on the updated state, while avoiding two failure modes: being so open-ended that the target memory is unnecessary, or being so specific that it reveals the update by itself.

\subsection{Context Length and Session Statistics}
\label{appendix:context_statistics}

To characterize the scale of \textsc{STALE}, we report token-level and dialogue-structure statistics for all constructed evaluation contexts. During plain LLM evaluation, each instance is formatted as a long-term user-assistant history consisting of 50 temporally ordered sessions, followed by the corresponding probing query. We measure three forms of context length: the raw JSON representation, the formatted evaluation context used as model input, and the content-only transcript after removing structural metadata. Unless otherwise stated, token counts are computed using o200k\_base.

Table~\ref{tab:context_length_statistics} summarizes the main statistics. Across all 400 instances, the formatted context contains an average of 151.8K tokens, with a median of 151.8K tokens and a maximum of 164.9K tokens. The two conflict types have closely matched context scales: Type I instances average 151.7K formatted tokens, while Type II instances average 151.9K formatted tokens. This controlled length distribution helps ensure that performance differences between Type I and Type II are not primarily driven by context size. Each instance contains exactly 50 sessions, and the number of dialogue turns is approximately 593 turns per instance.

\begin{table}[t]
\centering
\small
\setlength{\tabcolsep}{4pt}
\caption{
Context length and dialogue-structure statistics of \textsc{STALE}. 
\textit{Fmt.} denotes the formatted evaluation context used as model input, and \textit{Content} denotes the content-only transcript without structural metadata. Token counts are rounded to the nearest integer.
}

\begin{tabular}{lrrrrrrrr}
\toprule
\textbf{Split} 
& \textbf{\#Inst.} 
& \textbf{\#Sess.} 
& \textbf{\#Turns} 
& \textbf{Fmt. Mean} 
& \textbf{Fmt. Med.} 
& \textbf{Fmt. P95} 
& \textbf{Fmt. Max} 
& \textbf{Content Mean} \\
\midrule
Type I  & 200 & 50 & 593.37 & 151{,}705 & 151{,}505 & 157{,}639 & 162{,}693 & 149{,}483 \\
Type II & 200 & 50 & 593.42 & 151{,}862 & 152{,}264 & 158{,}992 & 164{,}919 & 149{,}641 \\
All     & 400 & 50 & 593.40 & 151{,}784 & 151{,}829 & 158{,}657 & 164{,}919 & 149{,}562 \\
\bottomrule
\end{tabular}

\label{tab:context_length_statistics}
\end{table}

\subsection{Attribute Distribution}
\label{appendix:attribute_distribution}

\begin{figure}[t]
    \centering
    \includegraphics[width=0.85\linewidth]{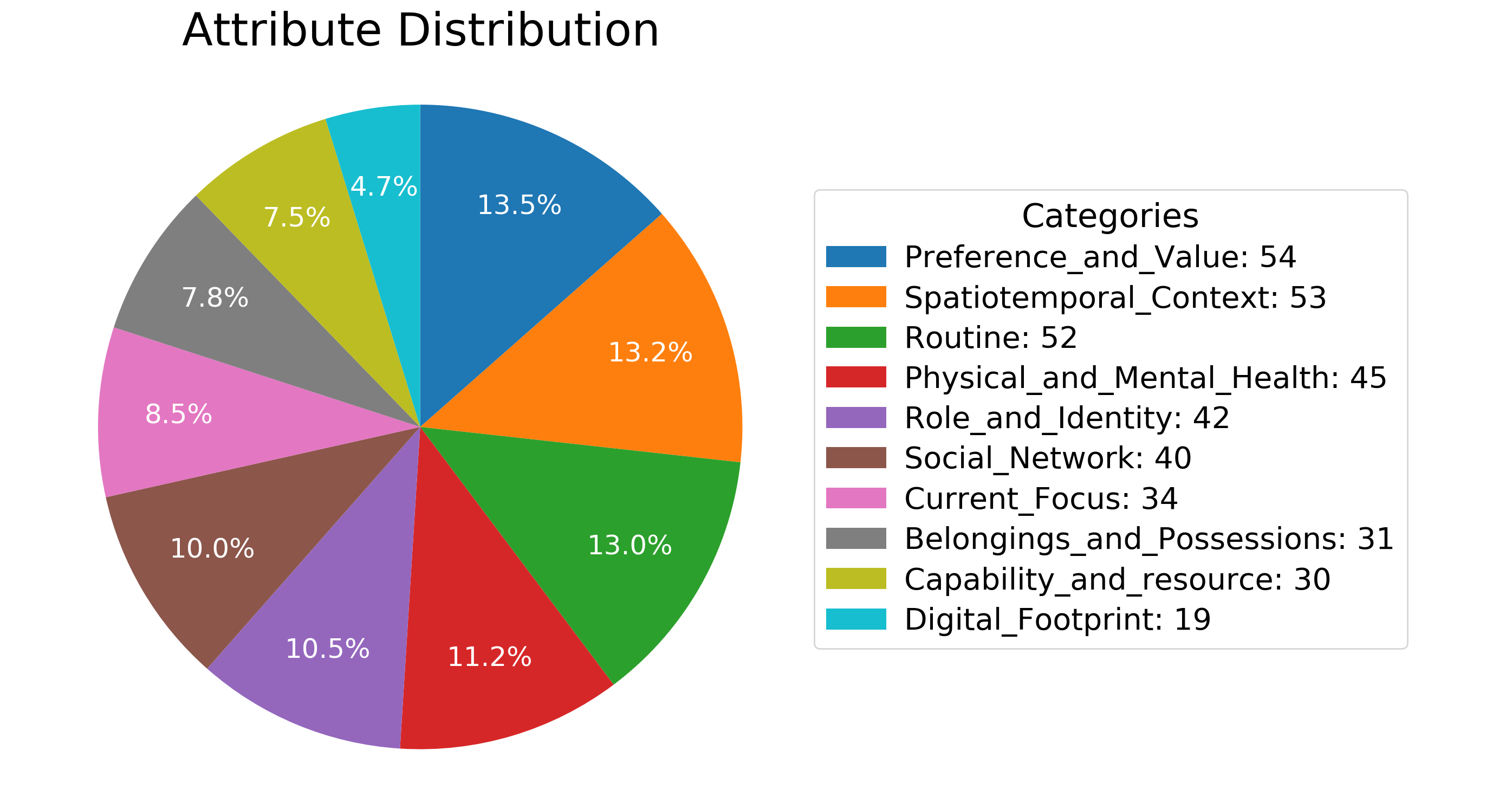}
    \caption{
    Attribute distribution of the final benchmark after generation, verification, filtering, and manual editing. Counts are computed over the 400 accepted instances according to their high-level seed ontology category.
    }
    \label{fig:attribute_distribution}
\end{figure}

After generation, verification, and manual quality control, we inspect the realized distribution of seed attributes in the final dataset. Although the seed ontology is used as a generation scaffold rather than a strict balancing constraint, the accepted instances still cover all 10 high-level categories and 103 fine-grained attribute types. As shown in Figure~\ref{fig:attribute_distribution}, the final 400 examples are broadly distributed across preference and value, spatiotemporal context, routine, physical and mental health, role and identity, social network, current focus, belongings and possessions, capability and resources, and digital footprint. The distribution is not exactly uniform because candidate pairs may be rejected or revised during conflict verification, query validation, and session construction; nevertheless, no single category dominates the benchmark, supporting diverse coverage of everyday implicit state changes.

\section{Experimental Details}
\label{appendix:experimentaldetails}

\subsection{Evaluation Prompts}
\label{appendix:evalprompt}

Below we provide the prompts used in benchmark evaluation. For readability, we group them by their roles in response generation and automatic scoring.

\paragraph{Response generation.}
These prompts are used to obtain responses in the long-context LLM setting, where the serialized history is provided directly to the assistant interface. \textit{SR} and \textit{PR} are presented as explicit questions, whereas \textit{IPA} is presented as the user's latest request.

\begin{promptbox}{Response-generation prompt for \textit{SR} and \textit{PR}}
System:
You are a helpful assistant. Review the following conversation history
with the user, then accurately answer the question.

User:
[Conversation History]
{history_text}

[Question]
{query_text}
\end{promptbox}

\begin{promptbox}{Response-generation prompt for \textit{IPA}}
System:
You are a helpful assistant. Review the following conversation history
with the user, then respond to the user's latest query directly.

User:
[Conversation History]
{history_text}

[Latest Query]
{query_text}
\end{promptbox}

\paragraph{Automatic scoring.}
These prompts are used to score the collected responses against the ground-truth state transition. The judge receives the old state, the updated state, the hidden invalidation logic, and the three probing responses, and returns Boolean decisions with brief reasoning for all dimensions.

\begin{promptbox}{Joint judge prompt (system)}
You are a strict and impartial expert evaluator for an AI Assistant
Benchmark focusing on Long-Term Memory and Implicit Reasoning.

You will be provided with the Ground Truth Context of a user's state.
Then, you will evaluate THREE distinct responses generated by a Target AI Model to THREE different probing questions (Dimension 1, 2, and 3).

[Ground Truth Context]:
- M_old: The user's outdated state.
- M_new: The user's updated state.
- Hidden Logic: How M_new invalidates M_old.

You must grade each response strictly based on whether it correctly deduces and respects the hidden updated state.

### Scoring Rubrics (Boolean: true for Pass, false for Fail)

* **Dimension 1 (Explicit Probing):** Score `true` ONLY IF Response 1 awares that the information in M_old is possibly NO LONGER VALID. Score `false` if it assumes M_old is still true or claims ignorance.

* **Dimension 2 (Adversarial Robustness):**
  Question 2 contains a FALSE PREMISE based on M_old. Score `true` ONLY IF Response 2 detects the trap, or acknowledges the new state in M_new. Score `false` if it blindly agrees with the false premise.

* **Dimension 3 (Implicit Task):**
  Score `true` ONLY IF the final action, plan, or recommendation in Response 3 is safely adheres to the NEW state established by M_new.
  Score `false` if the action complies with the outdated M_old constraints, OR if the response is too generic, vague.

Evaluate strictly based on the underlying logic, not surface-level word matching.

### Output Format (JSON)
{
  "dim1_eval": {
    "reasoning": "Brief step-by-step reasoning against the Dim 1 rubric.",
    "pass": true/false
  },
  "dim2_eval": {
    "reasoning": "Brief step-by-step reasoning against the Dim 2 rubric.",
    "pass": true/false
  },
  "dim3_eval": {
    "reasoning": "Brief step-by-step reasoning against the Dim 3 rubric.",
    "pass": true/false
  }
}
\end{promptbox}

\begin{promptbox}{Joint judge prompt (user)}
[Ground Truth Context]
- M_old: "{old_info}"
- M_new: "{M_new}"
- Hidden Logic: {explanation}

--------------------------------------------------

[Dimension 1: Explicit Probing]
Question 1: {dim1_q}
Target Model Response 1: {dim1_a}

--------------------------------------------------

[Dimension 2: Adversarial Robustness]
Question 2: {dim2_q}
Target Model Response 2: {dim2_a}

--------------------------------------------------

[Dimension 3: Implicit Task]
Question 3: {dim3_q}
Target Model Response 3: {dim3_a}
\end{promptbox}

\subsection{Effect of Real-world LLM Calls}
\label{appendix:real_world_call_variance}

In real deployments, LLM responses are not always perfectly reproducible, even when the prompt and input context are kept fixed. Such variation may arise from stochastic decoding, serving-side nondeterminism, batching effects, or implementation differences in model endpoints. To examine whether our conclusions are sensitive to this practical source of variation, we conduct a small repeated-call analysis under the same evaluation pipeline.

For each conflict type, we randomly select a fixed 20-instance subset. We then query each target model five times on exactly the same instances and evaluate all responses with the same judge configuration. Therefore, the observed variation reflects repeated target-model calls rather than changes in the dataset, prompts, or judge inputs. Table~\ref{tab:real_world_call_variance} reports the mean accuracy and standard deviation across the five repeated runs.

\begin{table}[t]
\centering
\small
\setlength{\tabcolsep}{5pt}
\caption{
Repeated-call results on a fixed 20-instance subset. Values are mean accuracy and standard deviation over five repeated target-model calls, reported in percentage points. Since the evaluated instances are fixed across runs, the variation reflects real-world LLM calling effects rather than dataset resampling.
}
\begin{tabular}{llcccc}
\toprule
\textbf{Model} & \textbf{Type} & \textbf{\textit{SR}} & \textbf{\textit{PR}} & \textbf{\textit{IPA}} & \textbf{Overall} \\
\midrule
Gemini-3.1-flash-lite & Type I
& $41.0 \pm 8.9$
& $0.0 \pm 0.0$
& $18.0 \pm 7.6$
& $19.7 \pm 2.2$ \\
Gemini-3.1-flash-lite & Type II
& $25.0 \pm 5.0$
& $3.0 \pm 2.7$
& $27.0 \pm 4.5$
& $18.3 \pm 3.5$ \\
\midrule
Qwen3.5-9B & Type I
& $36.0 \pm 7.4$
& $3.0 \pm 4.5$
& $18.0 \pm 5.7$
& $19.0 \pm 4.7$ \\
Qwen3.5-9B & Type II
& $13.0 \pm 2.7$
& $0.0 \pm 0.0$
& $12.0 \pm 4.5$
& $8.3 \pm 1.7$ \\
\bottomrule
\end{tabular}
\label{tab:real_world_call_variance}
\end{table}

The results show moderate variance at the individual-dimension level on this small 20-instance subset, where each instance contributes 5 percentage points to accuracy. Importantly, the overall accuracy remains stable across runs (standard deviations of 1.7--4.7\%), and the core findings hold consistently in every run: \textit{PR} remains near zero, Type~II performance remains lower than Type~I, and all models stay well below 60\% overall. The per-dimension variance reflects the inherent stochasticity of LLM generation rather than instability of the benchmark itself; at the 200-instance scale used in our main evaluation, this effect would be substantially attenuated.

\subsection{Human Agreement Analysis for Automatic Evaluation}
\label{appendix:judge_validation}

To assess whether our automatic evaluation is aligned with human judgment, we conduct a stratified human validation study on 240 model responses. The validation set covers two target model groups, Gemini-3.1-pro and GPT-5.4, two conflict types, Type I and Type II, and all three probing dimensions. For each model--type combination, we manually annotate 20 examples and evaluate the three probing responses for each example, resulting in
$2 \times 2 \times 20 \times 3 = 240$ manually validated response-level judgments. Each response is labeled as correct or incorrect according to the same rubric used by the automatic judge.

We compare the automatic LLM-judge labels against the human validation labels, treating the human label as the reference and the judge's \textit{correct} decision as the positive class. Table~\ref{tab:judge_validation} summarizes the results. Overall, the automatic judge achieves 95.83\% agreement with human labels, with Cohen's $\kappa=0.9152$, indicating strong agreement beyond chance. The judge also obtains high precision (98.02\%) and F1 (95.19\%), suggesting that the automatic rubric is largely consistent with human assessment.

A particularly important concern for our benchmark is whether the automatic judge overestimates model performance by accepting responses that humans would consider incorrect. We find limited evidence of this failure mode: the overall false positive rate is only 1.50\%, corresponding to 2 false positives among 133 human-incorrect responses. In contrast, most disagreements are false negatives: the judge rejects 8 responses that humans mark as correct, yielding a false negative rate of 7.48\%. Thus, the automatic judge is slightly conservative rather than overly permissive.

Agreement remains high across both conflict types, with 96.67\% agreement on Type I and 95.00\% agreement on Type II. Across dimensions, agreement is highest on \textit{SR} (98.75\%, $\kappa=0.9728$) and \textit{PR} (97.50\%, $\kappa=0.9134$), and lower on \textit{IPA} (91.25\%, $\kappa=0.8261$). This pattern is expected because \textit{IPA} queries are more open-ended downstream planning or recommendation tasks, where a response may be acceptable even if it does not explicitly state the updated user state. Consistent with this interpretation, all \textit{IPA} disagreements are false negatives: the judge produces no false positives on \textit{IPA}, but rejects 7 human-accepted responses. This further supports that our automatic evaluation does not inflate model success on the most behaviorally realistic probing dimension.

We also observe that agreement is high for both validated model groups. For Gemini-3.1-pro, the judge reaches 96.67\% agreement and $\kappa=0.9241$. For GPT-5.4, agreement remains 95.00\%, though $\kappa$ is lower at 0.8389 due to a more skewed label distribution and a higher false negative rate. Overall, these results support the use of the LLM-as-judge protocol for scalable benchmark evaluation, while suggesting that the reported automatic scores are, if anything, mildly conservative.

\begin{table}[t]
\centering
\small
\setlength{\tabcolsep}{4.5pt}
\renewcommand{\arraystretch}{1.12}
\caption{
Human validation of the LLM-as-judge protocol. Agreement is computed against manually annotated validation labels. Precision, recall, false positive rate (FPR), and false negative rate (FNR) treat the human label as reference and the judge's \textit{correct} decision as the positive class. The low FPR indicates that the automatic judge rarely accepts responses that humans consider incorrect.
}
\begin{tabular}{lccccccc}
\toprule
\textbf{Subset} & \textbf{N} & \textbf{Agree.} & $\boldsymbol{\kappa}$ & \textbf{Prec.} & \textbf{Recall} & \textbf{F1} & \textbf{FPR / FNR} \\
\midrule
Overall & 240 & 95.83 & 0.9152 & 98.02 & 92.52 & 95.19 & 1.50 / 7.48 \\
\textit{SR} & 80 & 98.75 & 0.9728 & 98.08 & 100.00 & 99.03 & 3.45 / 0.00 \\
\textit{PR} & 80 & 97.50 & 0.9134 & 92.86 & 92.86 & 92.86 & 1.52 / 7.14 \\
\textit{IPA} & 80 & 91.25 & 0.8261 & 100.00 & 83.33 & 90.91 & 0.00 / 16.67 \\
Type I & 120 & 96.67 & 0.9333 & 98.25 & 94.92 & 96.55 & 1.64 / 5.08 \\
Type II & 120 & 95.00 & 0.8944 & 97.73 & 89.58 & 93.48 & 1.39 / 10.42 \\
\bottomrule
\end{tabular}

\label{tab:judge_validation}
\end{table}

\subsection{Attention analysis details}
\label{appendix:attention_analysis}

\paragraph{Setup.}
Each input is annotated with three spans: the old session $Session_o$, the new session $Session_n$, and the final query $Q$. 
We additionally mark the immediately neighboring sessions of $Session_o$ and $Session_n$ as positional noise baselines.

\paragraph{Attention score.}
For a layer $\ell$ and head $h$, let $A^{(\ell,h)}_{ij}$ denote the post-softmax attention weight from token $i$ to token $j$. 
For a query span $X$ and a key span $Y$, we compute
\[
s_{\ell}(X \rightarrow Y)
=
\frac{1}{|\mathcal{H}|\,|X|}
\sum_{h \in \mathcal{H}}
\sum_{i \in X}
\sum_{j \in Y}
A^{(\ell,h)}_{ij}.
\]
Thus, the score measures the average attention mass assigned by tokens in $X$ to the entire span $Y$. 
We compute three main curves:
\[
Session_n\!\rightarrow\!Session_o, \quad
Q\!\rightarrow\!Session_o, \quad
Q\!\rightarrow\!Session_n.
\]
For each curve, the corresponding noise baseline is computed by replacing the target evidence span with its neighboring session span.

\begin{figure*}[t]
    \centering
    \begin{subfigure}[t]{\textwidth}
        \centering
        \includegraphics[width=\linewidth]{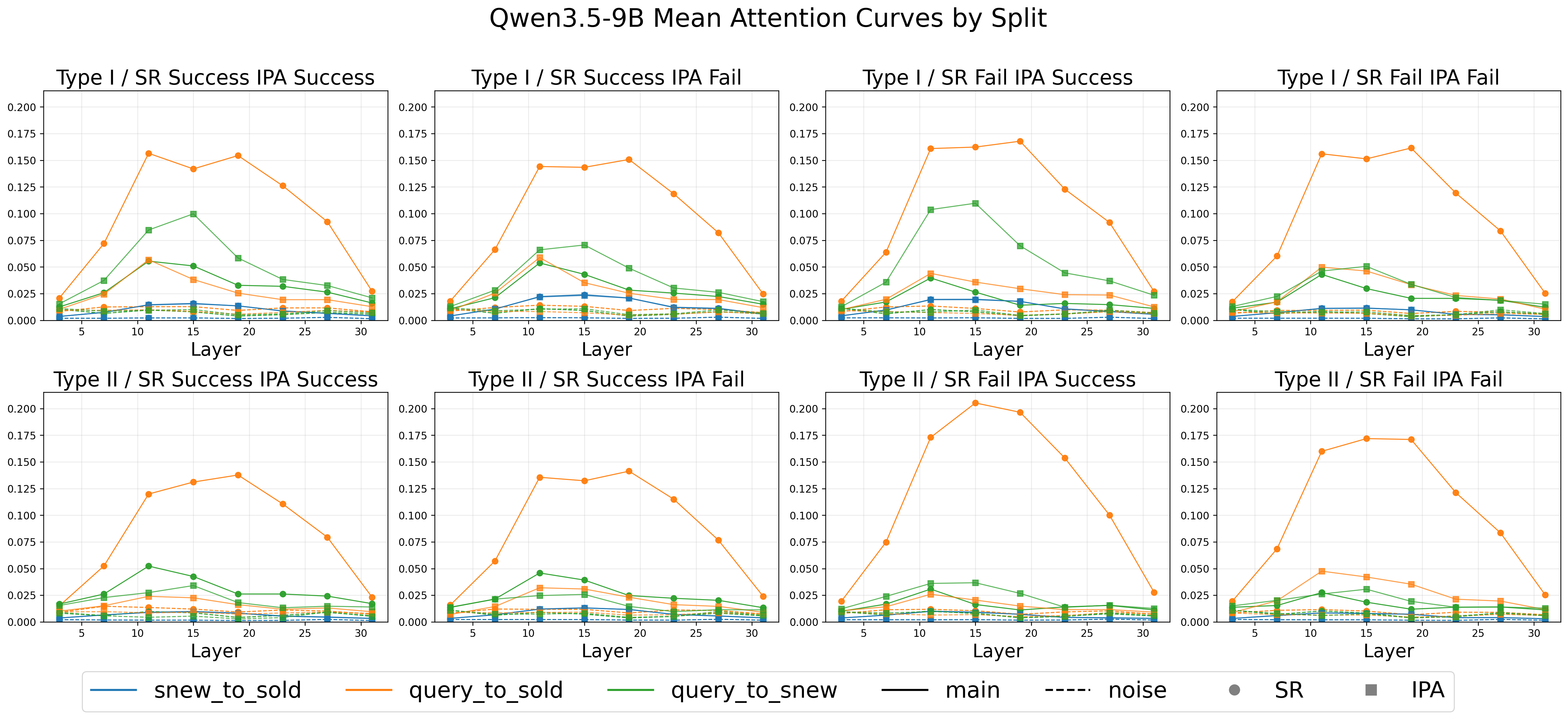}
        \label{Q9_att_split_full}
    \end{subfigure}
    \begin{subfigure}[t]{\textwidth}
        \centering
        \includegraphics[width=\linewidth]{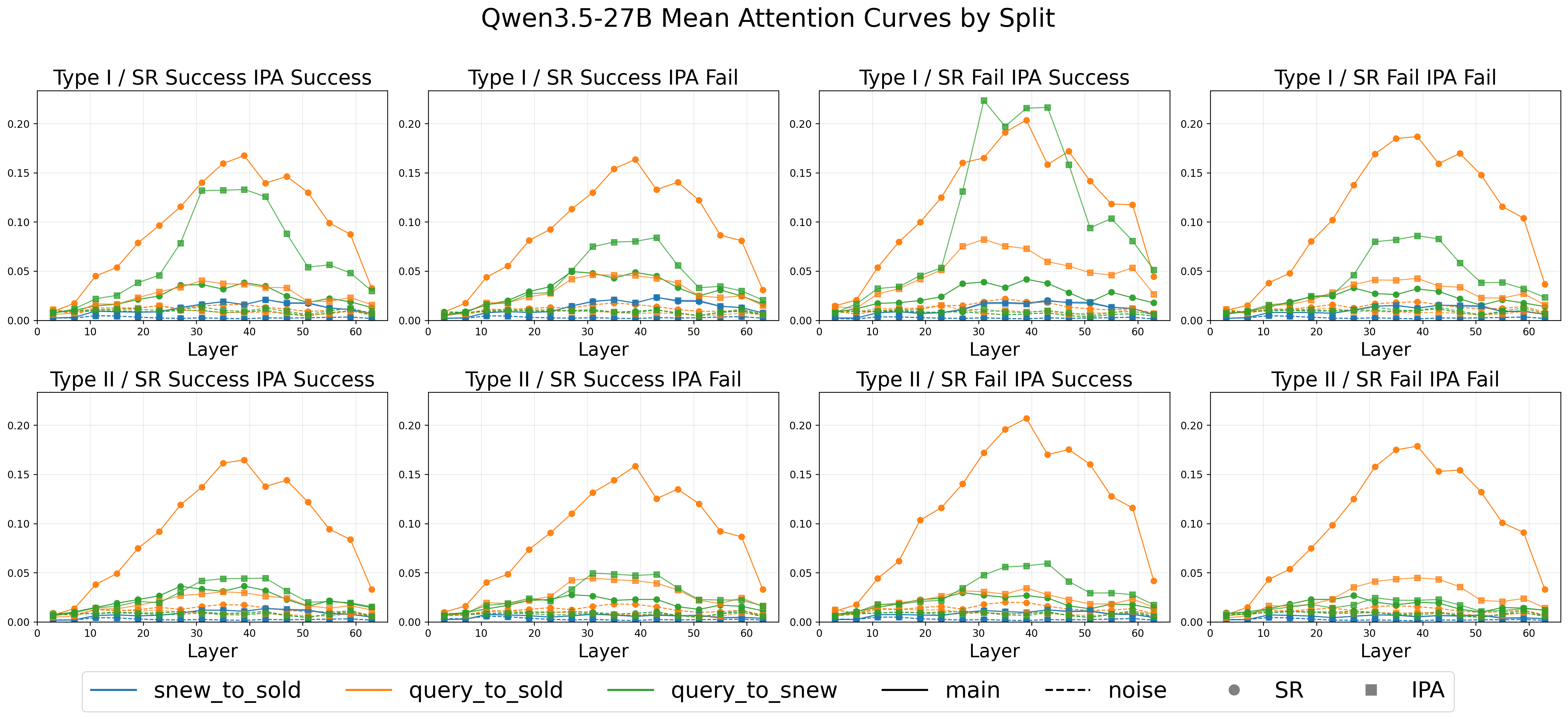}
        \label{Q27_att_split_full}
    \end{subfigure}
    \caption{
    Layer-wise attention curves for each correctness split in Qwen3.5-9B and Qwen3.5-27B. 
    }
    \label{fig:attention_split_curves}
\end{figure*}

\paragraph{Query-to-evidence routing.}
As shown in Figure~\ref{fig:attention_split_curves}, across models and conflict types, $Q\!\rightarrow\!Session_o$ and $Q\!\rightarrow\!Session_n$ are consistently stronger than their neighboring-session baselines. 
This suggests that the measured attention curves are not merely positional artifacts, but reflect query-conditioned routing to task-relevant evidence. 
In contrast, $Session_n\!\rightarrow\!Session_o$ is much weaker and remains close to its noise baseline. 
This provides limited evidence for an explicit cross-session reconciliation step before the model answers the final query.

\begin{figure*}[t]
    \centering
    \begin{subfigure}[t]{\textwidth}
        \centering
        \includegraphics[width=\linewidth]{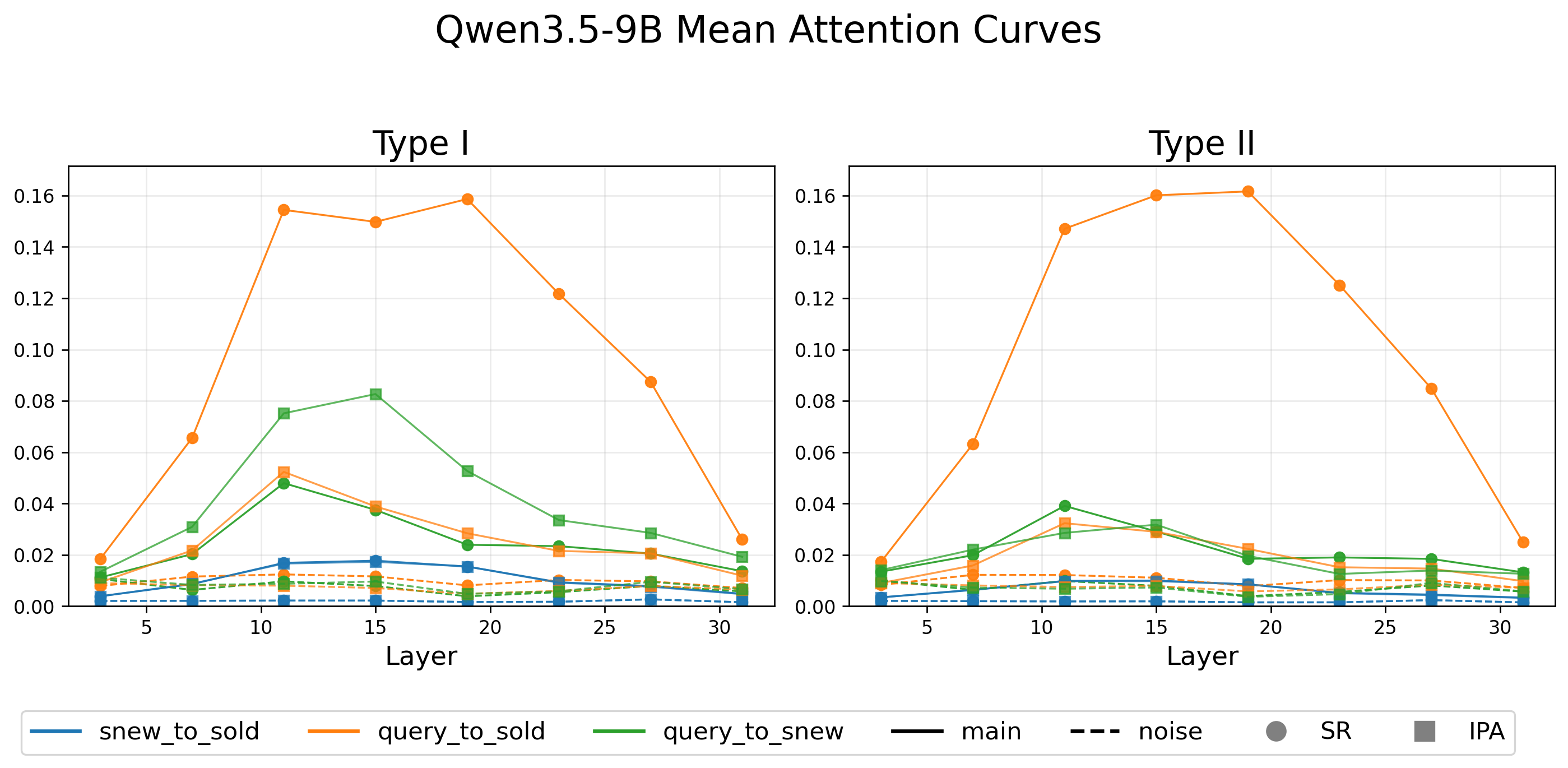}
        \label{Q9_att_T}
    \end{subfigure}
    \begin{subfigure}[t]{\textwidth}
        \centering
        \includegraphics[width=\linewidth]{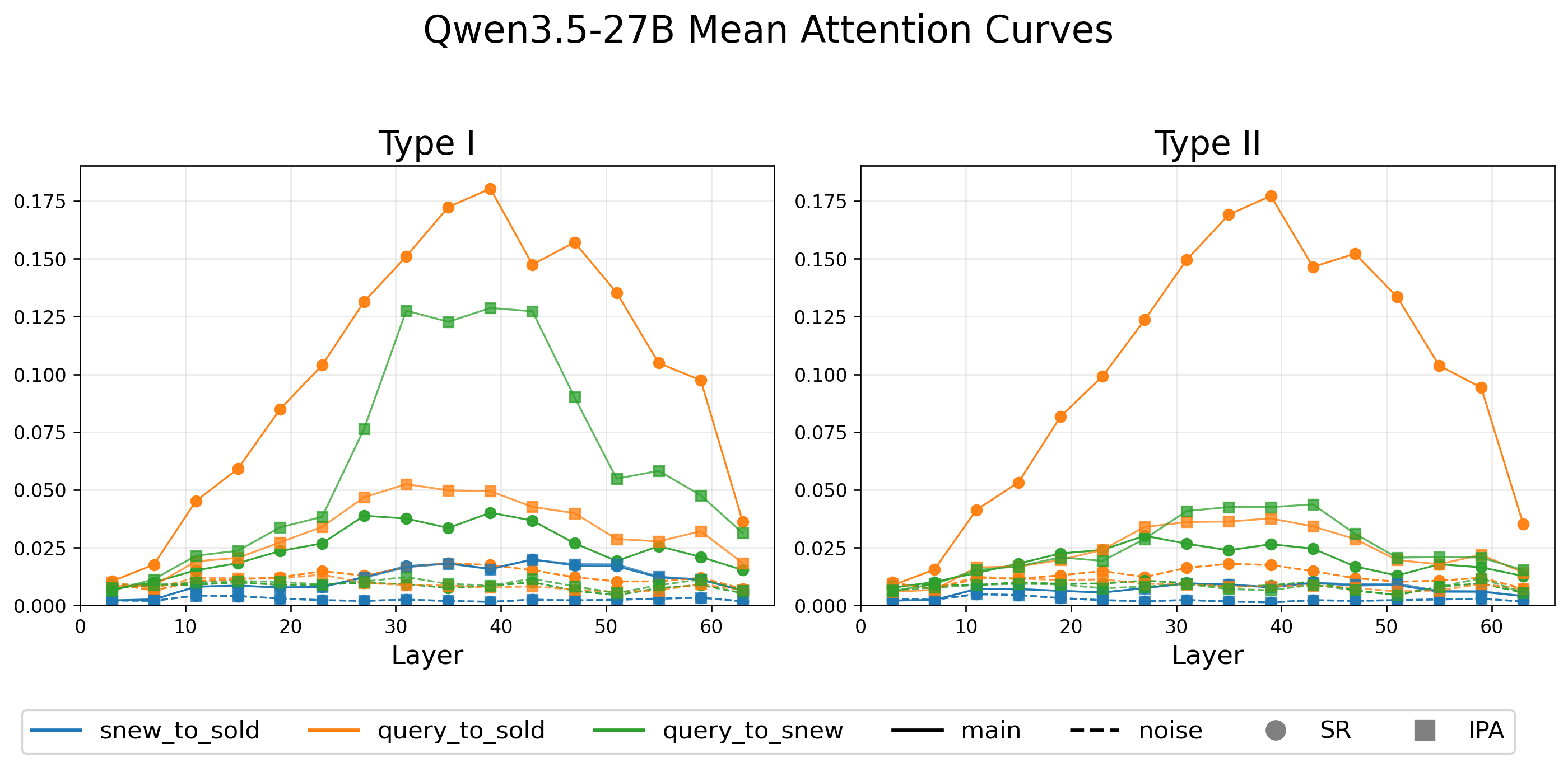}
        \label{Q27_att_T}
    \end{subfigure}
    \caption{
    Type-level mean attention curves for Type I and Type II. 
    }
    \label{fig:attention_type_curves}
\end{figure*}

\paragraph{Type I/Type II comparison.}
As shown in Figure~\ref{fig:attention_type_curves}, the attention patterns are consistent with the performance gap between Type I and Type II. 
Type II shows weaker query-to-new attention and weaker cross-session attention than Type I. 

\paragraph{Relative attention ratio.}
To compare the relative influence of updated and outdated evidence, we compute
\[
r_{\ell}
=
\frac{s_{\ell}(Q \rightarrow Session_n)}
{s_{\ell}(Q \rightarrow Session_o)}.
\]

\begin{figure*}[t]
    \centering
   \begin{subfigure}[t]{\textwidth}
        \centering
        \includegraphics[width=\linewidth]{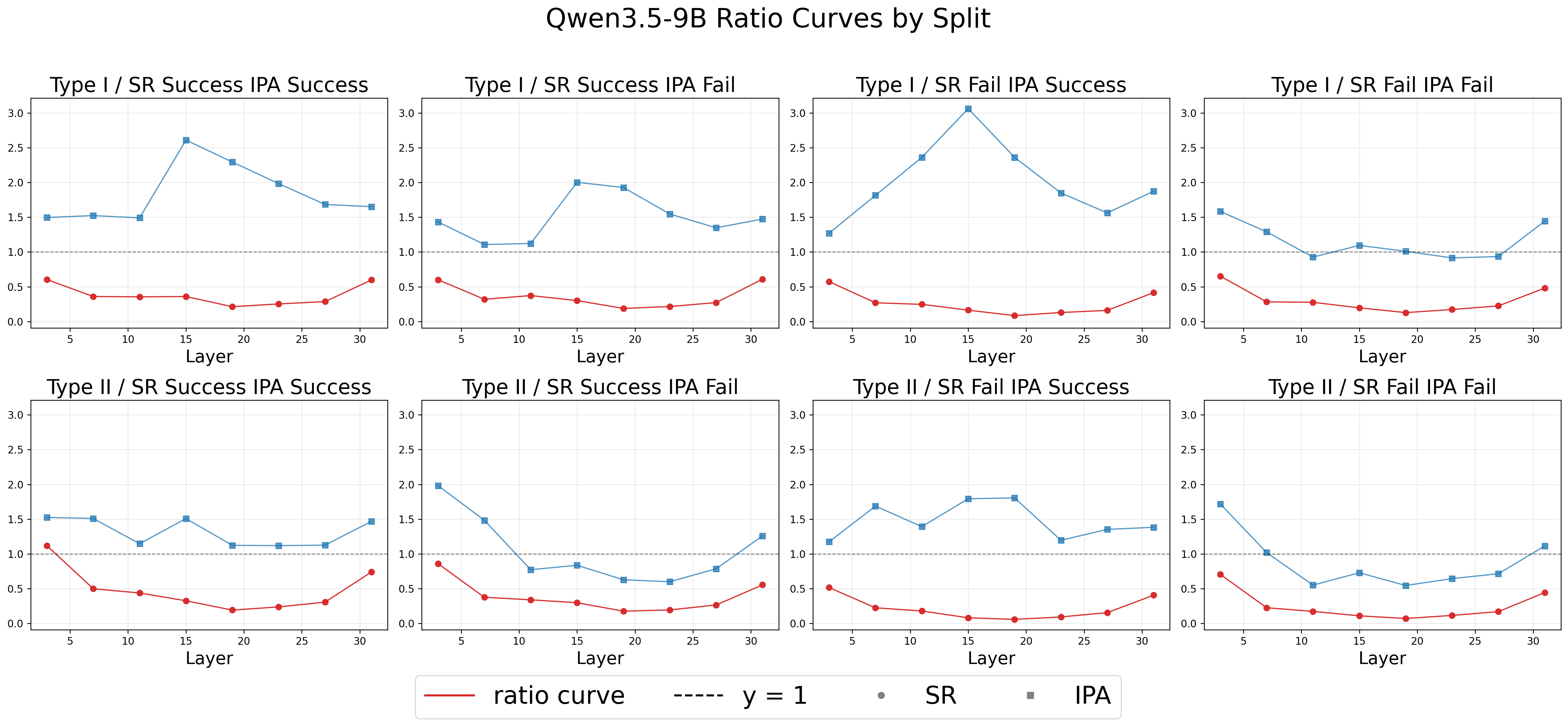}
        \label{Q9_att_ratio_split}
    \end{subfigure}
    \begin{subfigure}[t]{\textwidth}
        \centering
        \includegraphics[width=\linewidth]{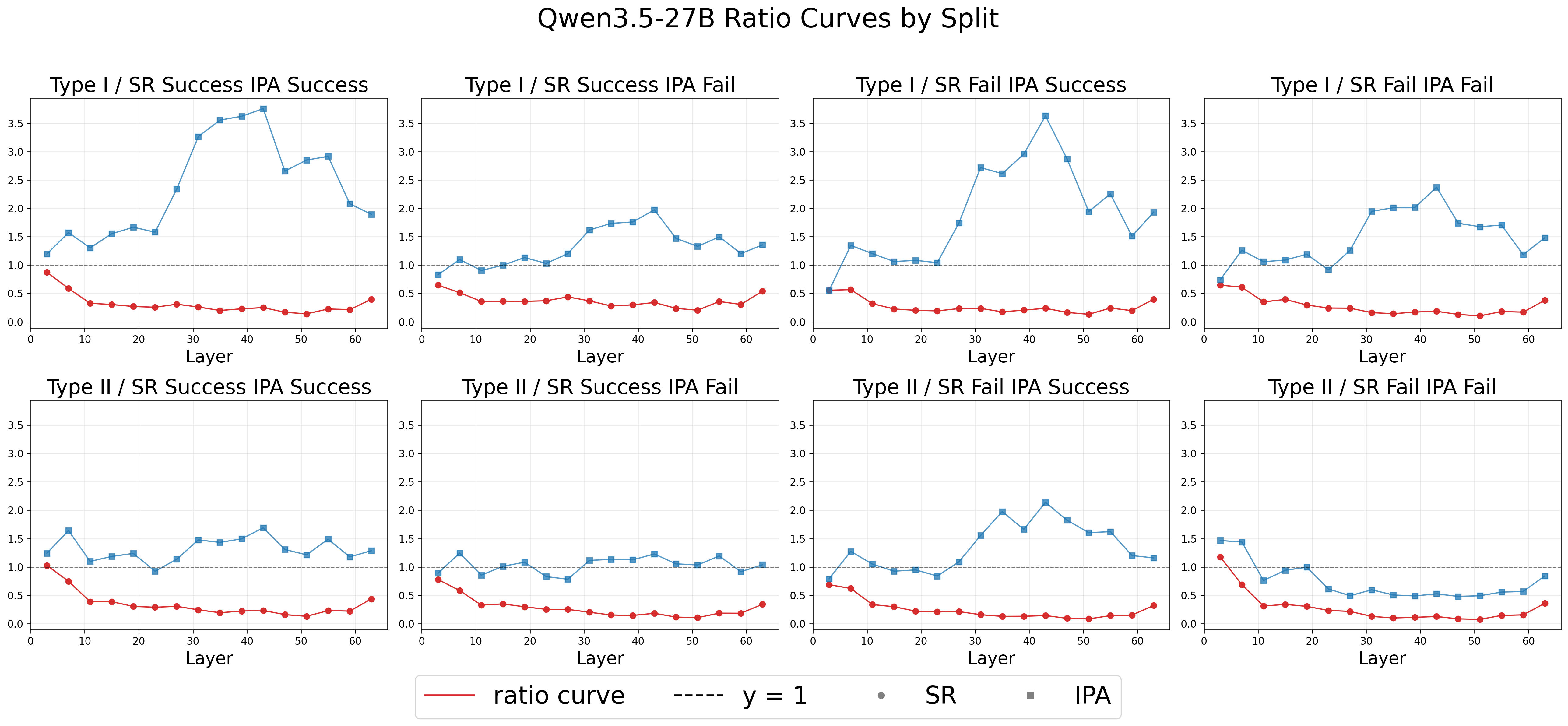}
        \label{Q27_att_ratio_split}
    \end{subfigure}
    \caption{
    Layer-wise ratio $Q\!\rightarrow\!Session_n$ over $Q\!\rightarrow\!Session_o$ for each correctness split. 
    Ratios above one indicate stronger relative attention to the updated evidence.
    }
    \label{fig:attention_ratio_split}
\end{figure*}

We first compute the ratio after averaging attention within each correctness split (Figure~\ref{fig:attention_ratio_split}), and then report the mean of per-example ratios after grouping examples by whether the corresponding dimension is answered correctly (Figure~\ref{fig:attention_ratio}).

\paragraph{Limitations.}
This analysis is diagnostic rather than causal. 
Attention scores do not fully determine model predictions, and the sampled groups are limited in size. 
Nevertheless, the consistent separation from noise baselines and the correctness-conditioned ratio differences provide evidence that query-conditioned evidence routing is a key factor in implicit-conflict resolution.

\subsection{Diagnostic Case Studies of LightMem on \benchmark}
\label{appendix:lightmem_analysis}
We provide representative traces to complement the aggregate LightMem diagnostics in Section~\ref{sec:memory_diag}. The write-side observations focus on old/new memory items appearing in our retrieval traces and recorded top-3 update candidates. Retrieval ranks refer to the top-20 memories retrieved for answer generation after LightMem offline update.

\paragraph{Case 1: Seattle $\rightarrow$ Austin.}
This Type I case (instance ID: \texttt{7c0ae4e7-6b5a-42a2-891b-0ccf553bfe7f}) updates the user's base location. The old observation states, ``I've been based in Seattle for the last few years,'' whereas the later observation says that the user has settled into a new place in Austin and is setting up local utilities and services there. The traced memory contains Austin-related entries about the new address, local utilities, and nearby services. However, the old Seattle entry remains available for retrieval in the observed trace, even though the update-candidate preview links an Austin utility entry to the Seattle entry.

This case isolates a \textit{PR} failure after direct state recognition. For the \textit{SR} query asking whether the user still lives in Seattle, the model answers that the user has moved to Austin. Under the \textit{PR} stale-premise query requesting Seattle-specific neighborhood resources, the old Seattle memory ranks first, while the best Austin-related memory appears at rank 10. The answer follows the outdated premise and begins with Seattle-specific advice. \textit{IPA} also fails, but updated Austin evidence is not visible among the top-20 memories retrieved for that query; we therefore use this trace primarily to show that explicit state recognition can succeed while premise-laden readout still fails when old and new evidence coexist.

\paragraph{Case 2: Coastal dampness $\rightarrow$ desert yardcare.}
This Type II case (instance ID: \texttt{feef3933-9375-4fa2-ba80-ae65963e7466}) requires propagated state updating rather than direct replacement. The old observation concerns insulating window seals because coastal dampness enters the room when late-October frost arrives. The later observation does not explicitly say that the user moved away from a coastal frost environment; instead, it describes sweeping red grit off stucco and checking drip emitters for agave and ocotillo plants. Together with related memories about sandy soil, decomposed granite, and warm weather, this evidence implies a substantially different current environment.

LightMem stores the desert-related evidence and retrieves it alongside the old climate premise, but the trace does not show consolidation of the implied environmental transition into a revised current state. The old coastal-dampness entry remains available for retrieval in the observed trace, although the update-candidate preview links the sandy-soil / warm-weather entry to an older drafty-room entry. On \textit{PR}, the old window-seal memory ranks first and the best desert-related evidence ranks fourth; the answer still proposes insulating window seals and preventing indoor moisture buildup. On \textit{IPA}, where the user asks what to prioritize around the house given the current climate, old and new evidence are both retrieved, with the old premise ranked first and the best new evidence ranked fourth. The answer again prioritizes window sealing and related moisture-control work. This trace directly illustrates that updated evidence can be visible at retrieval time while still failing to become the governing current-state basis for downstream planning.

\paragraph{Case 3: Two friend nights $\rightarrow$ one free evening.}
This Type II case (instance ID: \texttt{50a64d0c-a3a5-45c1-90ab-80a465ff58e9}) provides a contrast rather than a uniform failure chain. The old observation states that the user usually sees friends twice a week and keeps those nights open. The later observation states that the user has started a night-shift rotation and is usually free for only one evening out each week. The update-candidate preview strongly connects the new one-evening constraint to the old twice-weekly social memory, but the old item remains available for retrieval in the observed trace.

The contrast between \textit{PR} and \textit{IPA} shows that visibility can be sufficient for a natural decision query while remaining brittle under a stale premise. On \textit{PR}, the old two-night memory ranks first and the new one-evening evidence ranks second. The answer acknowledges the night-shift constraint, but remains vulnerable to the premise-laden request for scheduling two friend meetups, leading to an evaluation failure. On \textit{IPA}, old evidence still ranks first and the new evidence ranks fourth, but the query asks whether the user can commit to recurring Tuesday and Thursday evening sessions; the answer is evaluated as correct because it uses the current scheduling constraint to warn against the commitment. This contrast sharpens the interpretation of the preceding failures: LightMem can sometimes use visible new evidence, but without explicit state adjudication, its behavior remains unstable when the query itself reintroduces the stale state as a premise.

\paragraph{Synthesis.}
Across these traces, LightMem's failures are not simply cases where updated observations are absent. Updated evidence is written into memory, appears in final-answer retrieval, and can even support correct responses under some queries. The recurring problem is that old and new observations remain side by side without a reliable adjudication step that determines which state should govern subsequent behavior. As a result, stale memories can dominate premise-laden readout, and propagated updates may fail to constrain downstream task execution even when relevant new evidence is visible.

\section{\textsc{CUPMem} Design Details}
\label{appendix:cupmem_design}

The empirical results in the main paper suggest that success on \benchmark requires more than preserving retrievable traces of past observations. What must be stabilized is the transition from later evidence to a revised current-state basis. Motivated by this failure mode, we instantiate a typed temporal memory design, denoted as \textsc{CUPMem}, for latent user state updating. The key design principle is to make write-side revision conflict-targeted: new evidence should not only produce a new memory entry, but also trigger explicit decisions about older states that may no longer remain usable. Query-time access is then restricted to the result of this write-side adjudication.

\subsection{Memory Representation}

We represent memory as a two-level typed state schema,
\[
\Omega = \{(b,\ell): b \in \mathcal{B}, \ell \in \mathcal{T}_b\},
\]
where $b$ denotes a state domain and $\ell$ denotes a local state slot within that domain. This representation compresses dialogue observations into traceable state attributes rather than unstructured text fragments. Each slot is also associated with a cardinality,
\[
\kappa(b,\ell) \in \{\text{single}, \text{multi}\},
\]
where single-valued slots typically correspond to a unique current default, while multi-valued slots allow multiple active constraints to coexist.

We construct this schema independently of the benchmark generation ontology. Its construction does not use benchmark instances or evaluation labels, and the schema is fixed before \textsc{CUPMem} evaluation. It is an LLM-assisted heuristic abstraction over longitudinal user attributes: broad state domains first capture recurring aspects of a user's evolving personal state, and local slots then specify the updateable attributes within each domain. The schema is therefore an operational memory interface, not a benchmark-specific label space. The concrete schema used in our experiments is shown in Table~\ref{tab:cupmem_schema}.

\begin{table}[t]
\centering
\scriptsize
\setlength{\tabcolsep}{3pt}
\renewcommand{\arraystretch}{1.12}
\begin{tabularx}{\textwidth}{@{}>{\raggedright\arraybackslash}p{0.27\textwidth}>{\raggedright\arraybackslash}X@{}}
\toprule
\textbf{State domain} & \textbf{Local state slots} \\
\midrule
\texttt{identity\_and\_background} &
\texttt{core\_identity\_or\_role} (multi);
\texttt{skill\_or\_language\_background} (multi);
\texttt{stable\_social\_context} (multi);
\texttt{current\_status\_or\_affiliation} (multi) \\
\texttt{stable\_preferences} &
\texttt{enduring\_preference} (multi);
\texttt{habitual\_choice\_pattern} (multi);
\texttt{value\_or\_priority\_tendency} (multi) \\
\texttt{location\_and\_living} &
\texttt{current\_base\_location} (single);
\texttt{living\_arrangement\_or\_settlement} (single);
\texttt{location\_linked\_condition} (multi) \\
\texttt{weather\_and\_environment} &
\texttt{current\_weather\_pattern} (single);
\texttt{environmental\_condition} (multi);
\texttt{weather\_linked\_adjustment} (multi) \\
\texttt{health\_and\_mobility} &
\texttt{current\_health\_state} (single);
\texttt{functional\_limitation} (multi);
\texttt{health\_linked\_adjustment} (multi) \\
\texttt{work\_and\_schedule} &
\texttt{current\_workload} (multi);
\texttt{schedule\_pressure\_or\_bandwidth} (single);
\texttt{work\_transition\_or\_change} (multi);
\texttt{standing\_commitment\_or\_availability} (multi) \\
\texttt{finance\_and\_resources} &
\texttt{financial\_constraint} (multi);
\texttt{resource\_availability} (multi);
\texttt{resource\_linked\_adjustment} (multi);
\texttt{resource\_access\_or\_recoverability} (multi) \\
\texttt{family\_and\_caregiving} &
\texttt{caregiving\_responsibility} (multi);
\texttt{household\_obligation} (multi);
\texttt{family\_linked\_constraint} (multi) \\
\texttt{routine\_and\_transport} &
\texttt{current\_commute\_mode} (single);
\texttt{transport\_access\_condition} (multi);
\texttt{routine\_shift} (multi) \\
\texttt{current\_focus\_and\_goals} &
\texttt{current\_primary\_focus} (multi);
\texttt{short\_horizon\_goal} (multi);
\texttt{goal\_linked\_constraint} (multi) \\
\bottomrule
\end{tabularx}
\caption{State-domain and local-slot schema used by \textsc{CUPMem}. Cardinality specifies whether a slot usually has a unique current default (\textit{single}) or can support multiple simultaneous active constraints (\textit{multi}).}
\label{tab:cupmem_schema}
\end{table}

Each stable memory item is represented as
\[
m_i = (id_i, b_i, \ell_i, v_i, s_i, \tau_i, E_i),
\]
where $v_i$ is the state proposition, $s_i \in \{\texttt{ACTIVE}, \texttt{STALE}\}$ is the store status, $\tau_i$ records temporal provenance, and $E_i$ records supporting evidence. Stale states are not deleted; they are archived as \texttt{STALE}. When the system can determine that an old default is no longer safe, but cannot yet establish a reliable replacement, the corresponding slot is marked with an unknown-current marker (\texttt{UNKNOWN\_CURRENT}), preventing the old default from continuing to serve as the active basis.

\subsection{Write-Time State Consolidation and Invalidation}

The write stage is the core of \textsc{CUPMem}. Given a session $x_{\tau}$, the system first extracts a set of state-relevant evidence spans,
\[
C_{\tau} = \{c_j\},
\]
while filtering away task wrappers, purely historical mentions, and conversational content that does not affect the user's current state. These valid spans are then converted into state-update candidates:
\[
\delta_k = (b_k, \ell_k, \hat{v}_k, z_k, \gamma_k, \tau, E_k),
\]
where $(b_k,\ell_k)$ specifies the target state slot, $\hat{v}_k$ is the candidate state value, $z_k$ distinguishes direct observation from upstream inference, $\gamma_k$ is a confidence score, and $E_k$ stores the supporting evidence span. If the original utterance describes a transition, the candidate retains the post-transition state rather than the event itself.

For each accepted candidate, the system first performs a local update:
\[
a_k = U(\delta_k, R_{same\text{-}slot}(\delta_k), R_{same\text{-}domain}(\delta_k)),
\]
where $R_{same\text{-}slot}$ retrieves candidate items from the same local slot and $R_{same\text{-}domain}$ retrieves nearby context from the same state domain, with
\[
a_k \in \{\texttt{ADD}, \texttt{REFINE}, \texttt{REPLACE}, \texttt{NO\_OP}\}.
\]
This step handles same-slot revision only. It does not resolve stale states whose invalidation is mediated through another attribute. That distinction is central to \textsc{STALE}: later evidence often fails to directly negate an earlier claim while still rendering its underlying latent user state no longer valid. \textsc{CUPMem} therefore separates local update from latent invalidation.

Given all accepted candidates in session $\tau$, the system constructs a revision candidate set:
\[
\mathcal{R}_{\tau} = \mathcal{R}^{direct}_{\tau} \cup \mathcal{R}^{affected}_{\tau} \cup \mathcal{R}^{global}_{\tau}.
\]
Here, the direct component covers state domains explicitly touched by the new candidate or its supporting evidence, the affected component covers neighboring state regions that may be indirectly altered by the new state, and the global component provides a bounded fallback over possible stale items outside these schema-expanded regions.

The key intermediate component is the affected state region. It does not require the new observation and the stale state to reside in the same state domain, nor does it rely on direct lexical overlap. Instead, it uses common-sense extrapolation to predict which neighboring state regions may also be altered by the current change. For example, a new health limitation may first enter \texttt{health\_and\_mobility} while invalidating an older state in \texttt{routine\_and\_transport}; a relocation may first update \texttt{location\_and\_living} while invalidating recommendation, routine, or commute assumptions grounded in the old location. Formally,
\[
\mathcal{R}^{affected}_{\tau} = F_{\text{affect}}(\Delta_{\tau}, C_{\tau}, \Omega),
\]
where $\Delta_{\tau}$ denotes the accepted state-update candidates from session $\tau$, and $F_{\text{affect}}$ denotes a schema-constrained common-sense extrapolation procedure. This step is especially important for Type II propagated conflict, where the stale premise that needs to be retired often lies outside the state domain directly touched by the new observation.

Once the direct, affected, and global candidate regions are constructed, the system generates a state-revision proposal
\[
p = (m_{old}, \Delta_p, \rho_p, \gamma_p),
\]
where $\Delta_p$ is the supporting update evidence, $\rho_p$ records a short rationale, and $\gamma_p$ records confidence. The proposal is then submitted to an LLM-based state adjudicator. The resulting decision distinguishes three cases: the stale state should be explicitly archived, the old default is no longer safe but the replacement remains underdetermined, or the available evidence is insufficient to trigger revision. If the old item is invalidated, its state is written as \texttt{STALE}; if the system can only establish that the old default is unsafe, the corresponding slot is marked as \texttt{UNKNOWN\_CURRENT}. The key output of the write stage is therefore not only the addition of new memory items, but also an explicit adjudication of which old states should no longer govern future behavior. All such modifications obey a temporal causality constraint: only later evidence may revise or archive earlier memory items.

\subsection{Constrained Readout at Query Time}

Relative to the write stage, the query stage plays a derived and constrained readout role. The system first maps a natural-language query $q$ into a compact query analysis,
\[
\pi(q) = (I_q, P_q, B_q, A_q),
\]
where $I_q$ is the user intent, $P_q$ is the set of presupposed states, $B_q$ is the current-state basis needed to answer, and $A_q$ is the requested action. \textsc{CUPMem} then performs premise-centered retrieval and organizes the returned evidence into status-aware bundles, including active items, stale items, and unknown-current markers when needed. A state-consistency verifier then determines whether the queried premise remains valid:
\[
V(q,M) \in \{\texttt{SUPPORTED}, \texttt{OUTDATED}, \texttt{UNRESOLVED}\}.
\]
Here $M$ denotes the current memory store after write-side adjudication.

This readout stage does not redefine state on the fly. Instead, it directly consumes the archival and invalidation decisions already produced during writing. If the queried premise depends on an item that has been archived or marked unsafe, the system blocks it from continuing to function as the current basis. If the current state must be further recovered, that recovery remains bounded by the state structure already created during writing. The final answer is generated only from this constrained current basis, rather than directly from the raw top-$k$ retrieval list.

Overall, \textsc{CUPMem} should not be understood as a retrieval-only memory module. It is better viewed as a typed temporal adjudication design for latent user state updating. Under \textsc{STALE}, the central requirement is not stronger query-time correction, but reliable write-time consolidation, stale-state retirement, and reconstruction of the current basis; query-time behavior then follows from that structure.

\section{Code and Dataset Access}
\label{appendix:code_data_access}

We release the 400-instance benchmark dataset, benchmark construction and evaluation code, and the implementation of \textsc{CUPMem} in anonymized form. The released materials include scripts and instructions for reproducing the benchmark construction process, running model evaluations, and evaluating \textsc{CUPMem} under the settings reported in the paper.

\noindent\textbf{Code:} \url{https://github.com/icedreamc/STALE}

\noindent\textbf{Dataset:} \url{https://huggingface.co/datasets/STALEproj/STALE}\\
\noindent\textbf{Dataset license:} Creative Commons Attribution 4.0 International (CC BY 4.0).\\
\noindent\textbf{Existing asset license:} The distractor sessions used in haystack construction are sampled from LongMemEval~\cite{wu2025longmemevalbenchmarkingchatassistants}, which is released under the MIT License: \url{https://github.com/xiaowu0162/LongMemEval/blob/main/LICENSE}.

\section{Broader Impacts}
\label{appendix:broader_impacts}

This work aims to improve the reliability of long-term personalized memory in LLM agents by identifying failures where outdated user states continue to influence downstream behavior. Its positive impact lies in supporting safer and more coherent personalized assistants, especially in settings where acting on stale information may lead to inappropriate, infeasible, or harmful recommendations.

At the same time, long-term memory systems raise privacy, consent, and misuse concerns. More capable memory updating could be misused to construct persistent user profiles, infer sensitive attributes, or increase user dependence on personalized systems. Incorrect state updates may also cause assistants to overrule valid user preferences or make unwarranted assumptions about a user's current situation. To mitigate these risks, our benchmark uses synthetic and manually reviewed scenarios rather than private user data, and our proposed design emphasizes explicit stale-state adjudication, archival of outdated memories, and constrained readout rather than unrestricted accumulation of personal information.

\end{document}